\documentclass[10pt,twocolumn,letterpaper]{article}

\usepackage[pagenumbers]{cvpr} % To force page numbers, e.g. for an arXiv version

\def\eqref#1{Eq.~(\ref{#1})}

\newcommand{\bb}[1]{{\mathbb{#1}}}
\newcommand*{\Scale}[2][4]{\scalebox{#1}{$#2$}}%

\def\bmzero{{\bm{0}}}

\def\bmmu{{\bm{\mu}}}

\newcommand{\bmI}{\bm{\mathrm{I}}}
\newcommand{\bmX}{\bm{\mathrm{X}}}

\newcommand{\bmx}{\bm{\mathrm{x}}}

\newcommand{\Var}{\mathrm{Var}}

\def\rmo{{\mathrm{co}}}

\def\rvepsilon{{\bm{\epsilon}}}

\def\gN{{\mathcal{N}}}

\usepackage{bm}
\usepackage{multirow}
\usepackage[accsupp]{axessibility}  % Improves PDF readability for those with disabilities.
\definecolor{cvprblue}{rgb}{0.21,0.49,0.74}
\usepackage[pagebackref,breaklinks,colorlinks,allcolors=cvprblue]{hyperref}

\def\paperID{13728} % *** Enter the Paper ID here
\def\confName{CVPR}
\def\confYear{2025}

\title{Unified Uncertainty-Aware Diffusion for Multi-Agent Trajectory Modeling}

\author{Guillem Capellera$^{1,2}$ \quad Antonio Rubio$^{2}$\quad Luis Ferraz$^{2}$\quad Antonio Agudo$^{1}$  \\  \textcolor{gray}{$^1$Institut de Robòtica i Informàtica Industrial, CSIC-UPC \quad  $^2$Kognia Sports Intelligence}  \\  {\tt\small \{guillem.capellera, antonio.agudo\}@upc.edu} \quad  {\tt\small \{antonio.rubio, luis.ferraz\}@kogniasports.com}}

\begin{document}
%\maketitle

%%%%%%%%%%%%%%%%%%%%%%  MAIN PAPER %%%%%%%%%%%%%%%%%%%%%%%%%%%%%
\twocolumn[{%
\renewcommand\twocolumn[1][]{#1} %
\maketitle
\thispagestyle{empty}
\begin{center}
    \centering
    %\vspace{2mm}
    \includegraphics[width=1.0\textwidth, trim={0 0.5cm 0 0}]{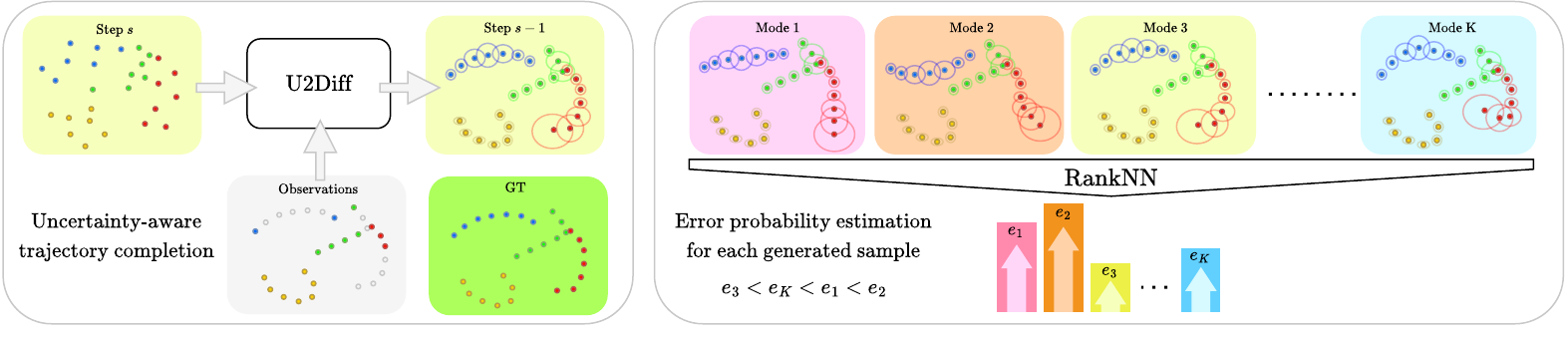}
    %\vspace{+0.2cm}
    \captionof{figure}{\textbf{Uncertainty-aware, unified and interpretable approach for trajectory modeling in multi-agent scenarios.} \textbf{U2Diff} is a diffusion-based model capable of performing trajectory completion tasks such as forecasting, imputation or inferring totally unseen agents, while also jointly estimating state-wise uncertainty. \textbf{RankNN} is a post-processing operation that infers an error probability for each generated mode under the same prior observations, which is strongly correlated with the error related to the ground truth.}
    \label{fig:teaser}
    %\vspace{-0.2cm}
\end{center}}]

\begin{abstract}
Multi-agent trajectory modeling has primarily focused on forecasting future states, often overlooking broader tasks like trajectory completion, which are crucial for real-world applications such as correcting tracking data. Existing methods also generally predict agents' states without offering any state-wise measure of uncertainty. Moreover, popular multi-modal sampling methods lack any error probability estimates for each generated scene under the same prior observations, making it difficult to rank the predictions during inference time. We introduce U2Diff, a \textbf{unified} diffusion model designed to handle trajectory completion while providing state-wise \textbf{uncertainty} estimates jointly. This uncertainty estimation is achieved by augmenting the simple denoising loss with the negative log-likelihood of the predicted noise and propagating latent space uncertainty to the real state space. Additionally, we incorporate a Rank Neural Network in post-processing to enable \textbf{error probability} estimation for each generated mode, demonstrating a strong correlation with the error relative to ground truth. Our method outperforms the state-of-the-art solutions in trajectory completion and forecasting across four challenging sports datasets (NBA, Basketball-U, Football-U, Soccer-U), highlighting the effectiveness of uncertainty and error probability estimation. 
%# UNCOMMENT HYPERLINK FOR ARXIV VERSION
\href{https://youtu.be/ngw4D4eJToE}{https://youtu.be/ngw4D4eJToE}
\end{abstract}

\section{Introduction}
Modeling trajectories in multi-agent settings is crucial for capturing stochastic human behaviors in various domains, including pedestrian motion prediction~\cite{alahi2016social,gupta2018social, amirian2019social, kosaraju2019social,salzmann2020trajectron++,ngiam2021scene,girgis2021latent, navarro2022social, saadatnejad2023social, xu2023eqmotion}, human pose estimation~\cite{fragkiadaki2015recurrent,jain2016structural,martinez2017human,mao2019learning,mao2020history,aksan2021spatio,cai2020learning,guo2023back}, and sports analytics~\cite{zheng2016generating, zhan2018generating, alcorn2021baller2vec++, hu2022entry, capellera2024footbots, peraltemporally, mao2023leapfrog, xu2023uncovering}. 

Multi-modal generative approaches primarily focus on forecasting future states based on past trajectories, utilizing models such as Generative Adversarial Networks (GANs)~\cite{dendorfer2021mg, gupta2018social}, Conditional Variational Auto-Encoders (CVAEs)~\cite{yuan2021agentformer,xu2022groupnet}, and, more recently, Denoising Diffusion Probabilistic Models (DDPM)~\cite{ho2020denoising}. DDPM have shown particular success in trajectory forecasting for applications like pedestrian and sports modeling~\cite{mao2023leapfrog, gu2022stochastic}. However, their evaluation is often limited to agent-wise metrics, overlooking scene-level dynamics that are crucial for multi-agent contexts. Additionally, these methods generally require fixed temporal window dimensions, which restricts their adaptability across diverse task settings and scenarios.

The task of \textit{trajectory completion} has emerged as a key advancement beyond traditional forecasting, enabling models to infer trajectories by leveraging both past and/or future observations~\cite{xu2023uncovering, liu2019naomi, qi2020imitative}. This task also seeks to predict totally unobserved agents using only the motions of the surrounding observable ones~\cite{kim2023ball, xu2025sportstraj, capellera2024transportmer}. This capability is especially relevant in sports, where complex multi-agent interactions require models to accurately capture both individual and coordinated team tactical behaviors within fixed spatial coordinates.

However, current state-of-the-art methods in both trajectory forecasting and completion focus primarily on predicting locations without estimating the uncertainty associated with each predicted state. This limitation highlights the need for a state-wise uncertainty estimation approach to quantify each state prediction’s closeness to the ground truth. Additionally, this gap presents a further challenge in developing methods to extract a scene-level uncertainty or error probability capable of ranking the reliability of multiple generated modes under the same prior.

In this study, we propose a \textbf{U}nified \textbf{U}ncertainty-aware \textbf{Diff}usion (U2Diff) aimed at tackling the general task of multi-agent trajectory completion while predicting per-state uncertainty with a novel variance propagation technique from latent to real space (see our pipeline in Fig.~\ref{fig:teaser}). Our method estimates global uncertainty by averaging the variances of each predicted agent's state. We show that this global uncertainty has certain correlation with the scene-level error across modes within the same prior, providing an unsupervised measure of confidence in the generated trajectories. To further refine the model’s interpretability, we propose a supervised Rank Neural Network (RankNN) in order to rank modes based on their proximity to ground truth values, providing error probabilities and achieving high Spearman correlation values, with medians around $0.58$ and $0.78$. 

We validate the effectiveness of our overall approach using four real-world sports datasets: two of basketball, one of football, and another of soccer; demonstrating substantial improvements over competing methods in scene-level metrics for forecasting and trajectory completion tasks. This work contributes a novel uncertainty-aware approach to trajectory modeling that enhances the reliability of generated trajectories in complex interactive environments like sports. The key contributions are summarized as:
\begin{itemize}
    \item We propose a diffusion-based approach for general trajectory completion in multi-agent domain, achieving state-of-the-art performance.
    \item We introduce a simple loss augmentation in diffusion framework that enables direct uncertainty estimation for each predicted state. It ensures consistency across timesteps and moderate correlation with ground truth error, while enhancing prediction robustness.
    \item We devise a post-processing supervised architecture (RankNN) providing error probability estimates for each generated mode under a shared prior, enabling high-correlation with ground truth error.
\end{itemize}

\section{Related Work}
We next review the most related work dealing with trajectory modeling, diffusion models and uncertainty estimation.

\noindent \textbf{Trajectory Modeling.} Multi-modal agent trajectory modeling has traditionally focused on predicting future positions from past observations. Early methods used Variational Recurrent Neural Networks (VRNNs) to capture stochasticity in human long term movement prediction~\cite{felsen2018will, zhan2018generating, zheng2016generating, sun2019stochastic, yeh2019diverse}. As the field evolved, GANs~\cite{dendorfer2021mg, gupta2018social,fang2020tpnet, hu2020collaborative, sadeghian2019sophie} and CVAEs~\cite{yuan2021agentformer, mangalam2020not, salzmann2020trajectron++, xu2022groupnet, lee2022muse} enabled more diverse and realistic future predictions by leveraging variational inference. Recently, diffusion models have demonstrated significant potential in generating diverse plausible futures~\cite{mao2023leapfrog, jiang2023motiondiffuser, gu2022stochastic, rempe2023trace, bae2024singulartrajectory, li2023bcdiff, yang2024diffusion}, surpassing previous methods in forecasting tasks. However, these approaches are often limited by fixed time horizons. Other methods, such as Graph Variational Neural Networks (GVRNNs)~\cite{xu2023uncovering, omidshafiei2022multiagent} and non-autoregressive techniques~\cite{liu2019naomi}, have been developed for trajectory imputation tasks. Building on these foundations, our work introduces a diffusion-based architecture that integrates forecasting and imputation in a unified framework, adaptable to multi-agent scenarios without predefined agent or time dimension constraints.

\noindent \textbf{Unified diffusion models.} Time-series diffusion models have emerged as a viable alternative to Gaussian processes for probabilistic modeling~\cite{tashiro2021csdi, alcaraz2022diffusion}. Our U2Diff architecture is inspired by CSDI~\cite{tashiro2021csdi}, which we adapted for multi-agent 2D trajectory modeling by employing a bidirectional version of MambaSSM~\cite{gu2023mamba} to enhance temporal processing, replacing Transformer Encoder~\cite{vaswani2017attention} focused on temporal dynamics processing. The sequential natural processing of MambaSSM allows to remove the temporal positional encoding while obtaining better performance. 

\noindent \textbf{Uncertainty-aware.} Traditional models predict positions but overlook state-level uncertainty. Recent work introduced global uncertainty measures by aggregating individual agent uncertainties~\cite{mao2023leapfrog}, but these lack the granularity needed to adapt diverse tasks such as trajectory completion. Inspired by pixel-wise uncertainty method in image generation~\cite{kou2023bayesdiff}, our model extends this to multi-agent trajectories, providing state-wise uncertainty, enabling finer-grained interpretability at state-level predictions.

\noindent \textbf{Probability estimation.} In multi-modal trajectory generation, probability estimation for each mode remains relatively unexplored. Existing methods assign probabilities using predefined trajectory anchors~\cite{chai2019multipath, phan2020covernet, shi2023trajectory} or post-process the predicted trajectory~\cite{zhao2021tnt}; however, they primarily focus on ego-agent scenarios. Latent sequential models~\cite{girgis2021latent, salzmann2020trajectron++} introduce the estimation of probabilities at the scene-level requiring a fixed number of modes. To address this and adapt to sampling-based approaches like U2Diff, we propose RankNN, which estimates error probabilities for each scene-mode using all agents' trajectories and uncertainties. Unlike prior methods, RankNN supports a variable number of modes under a shared prior, acting as a post-processing network which provides ranked error probability estimates.

%================================================
\section{Revisiting Diffusion Models}
\label{rection_revisiting}

We next review DDPM~\cite{ho2020denoising} that will be later employed to describe our method for uncertainty-aware multi-agent trajectory completion. They work by gradually adding random Gaussian noise to the original data in a {\em forward diffusion} process through a series of $S$ steps and then, learning to remove it in a {\em reverse denoising} one where original data is generated from the noise. To this end, let $\bmX_0$ be a data point from a real data distribution $q(\bmX)$ where $\bmX$ is the input data. Some Gaussian noise with variance $\beta_s\in (0,1)$ can be added to $\bmX_{s-1}$, obtaining a new latent variable $\bmX_{s}$ with distribution $q(\bmX_s \mid \bmX_{s-1})$ as:
\begin{align}
     q(\bmX_s \mid \bmX_{s-1}) &= \gN\left(\bmX_s; \sqrt{1-\beta_s} \bmX_{s-1}, \beta_s \bmI\right),\\
     q(\bmX_{1:S} \mid \bmX_0) &= \prod_{s=1}^{S} q(\bmX_s \mid \bmX_{s-1}),
    \label{eq:forward}
\end{align}
where $\bmI$ denotes an identity matrix, i.e., the distribution is always represented by a diagonal matrix of variances. Assuming a sufficiently large $S$, $\bmX_S \sim \gN(\bmzero,\bmI)$. $\bmX_s$ can be sampled at any arbitrary time step from the distribution: 
\begin{equation}
q(\bmX_s \mid \bmX_0) = \gN (\bmX_s; \sqrt{\hat{\alpha}_s}\bmX_0, (1-\hat{\alpha}_s)\bmI),
\end{equation}
where $\alpha_s=1-\beta_s$ and $\hat{\alpha}_s = \prod_{i=1}^s \alpha_i$. Then, $\bmX_s$ is expressed as:
\begin{equation}
\label{eq:sampling_way}
    \bmX_s = \sqrt{\hat{\alpha}_s}\bmX_0 + \sqrt{1-\hat{\alpha}_s} \rvepsilon, 
\end{equation}
where  $\rvepsilon \sim \gN(\bmzero,\bmI)$. Here $1 - \hat{\alpha_s}$ indicates the variance of the noise for an arbitrary time step, i.e., that could equivalently be used to define the noise schedule instead of $\beta_s$.

In the reverse diffusion process, a neural model is trained to infer the original data by reversing the previous noising process. Estimating $q(\bmX_{s-1} \mid \bmX_s)$ is a hard task as it depends on the entire data distribution and, therefore, a neural network $p_{\theta}(\cdot)$ is used to learn the $\theta$ diffusion parameters by parameterizing both mean and variance as:
\begin{align}
\small
     \label{eq:prior}
     p_\theta(\bmX_{s-1} \mid \bmX_s) &=   \gN(\bmX_{s-1}; \bmmu_\theta(\bmX_s,s),\sigma_\theta(\bmX_s,s)^2\bmI),\\
      p_\theta(\bmX_{0:S}) &= p(\bmX_S) \prod_{s=1}^{S} p_\theta(\bmX_{s-1} \mid \bmX_s), \label{eq:reverse1}
      %\label{eq:reverse}
\end{align}
%\vspace{-1mm}
where $p(\bmX_S) = \gN(\bmX_S; \bmzero,\bmI)$ and $\bmmu_\theta(\cdot)$ and $\sigma_\theta(\cdot)^2\bmI$ represent mean and covariance matrix, respectively. 

The mean in \eqref{eq:prior} can be obtained by considering the predicted noise $\rvepsilon_\theta (\bmX_s,s)$ at $s$ step as:
\begin{align}
  \bmmu_\theta(\bmX_s,s)&=\frac{1}{\sqrt{\alpha_s}} \left(\bmX_s-\frac{\beta_s}{\sqrt{1-\hat{\alpha}_s}} \rvepsilon_\theta (\bmX_s,s) \right),  \label{eq:mu} %\\ 
  %\sigma_\theta(\bmX_s,s) &= \tilde{\beta}_s^{1/2},
\end{align}
where $\rvepsilon_\theta$ is a trainable denoising function. In general, to infer the covariance matrix, the variance is assumed to be $\sigma_\theta(\bmX_s,s)^2 = \frac{1 - \hat{\alpha}_{s-1}}{1 - \hat{\alpha}_s} \beta_s$, i.e., it does not depend on the predicted noise. This parametrization is equivalent to rescaled score model for score-based generative models. Then, the reverse process can be trained by minimizing the function:
\begin{equation}
  \mathcal{L}_{\mathrm{simple}} = 
  \bb{E}_{\bmX_0 , \rvepsilon ,s}
  || \rvepsilon - \rvepsilon_{\theta}(\bmX_s, s) ||_2^2 ,
  \label{eq:loss}
\end{equation}
where $\rvepsilon$ is a random but known noise. Later, Nichol {\em et al.}~\cite{nichol2021improved} found that learning the variance $\sigma_\theta(\bmX_s,s)^2$ improved the log-likelihood in images, as we will do in this work.  

\section{Multi-Agent Trajectory modeling by diffusion models}

%Tashiro et al.\cite{tashiro2021csdi} proposed a diffusion probabilistic framework for handling multivariate time-series imputation, handling conditioning visible observations to infer the non-visible ones.

In this section we describe how to exploit probabilistic diffusion models to sort out trajectory completion in multi-agent scenarios. Our work is inspired by~\cite{tashiro2021csdi} that used a diffusion probabilistic framework for handling one-dimensional multivariate time-series imputation, exploiting visible observations to infer the non-visible ones. In contrast, in this work we present a unified approach for two-dimensional scenarios, where the relation between agents is richer and complex to capture.

\subsection{Problem Statement}
Let us consider a set of $N \in \mathbb{N}$ agent observations in a given time instant $t$, denoted as $\bmx_t = \{\bmx^n_t\}$ with $n=\{1,\ldots,N\}$, where each observation contains the $(x,y)$ locations. We can now collect $T$ observations along time for every agent, defining a scene tensor $\bmX$ where all $\bmx^{n}_{t}$ with $t=\{1,\ldots,T\}$ are considered. Trajectory completion aims at inferring missing or unobserved entries of a data structure based on the visible ones. Given a tensor of partial observations defined as $\bmX^\rmo$ and a $T \times N$ binary conditioning mask $\mathbf{M}$ to encode by 1 the visible observations and by 0 the unobserved ones, the goal is to find a function $f(\cdot)$ to infer the full observations such that:
\begin{equation}
\label{completion}
   \bmX=f(\bmX^\rmo, \mathbf{M} ).
\end{equation}

Particularly, in this paper we propose to model multi-agent trajectories by leveraging per-observation uncertainty estimation as:
\begin{equation}
\label{completion_distrib}
\Scale[0.95]{
   p(\bmX \mid \bmX^\rmo, \mathbf{M}) = \gN(\bmX; f^\mu(\bmX^\rmo, \mathbf{M}), f^{\sigma^2}(\bmX^\rmo, \mathbf{M})),}
\end{equation}
where $f^\mu(\cdot)$ and $f^{\sigma^2}(\cdot)$ denote the function to extract mean and covariance matrix, respectively. As we propose to employ a generative model to handle the previous problem, at inference time the method obtains $K\in \mathbb{N}$ modes or scenes according to the same prior observations such that:
\begin{equation}
\label{inference}
   p(\bmX^k \mid \bmX^\rmo, \mathbf{M}) \quad \forall k \in \{1, \ldots, K\}.
\end{equation}

Once the trajectory completion problem is addressed, we propose estimating an error probability for each mode, which must be correlated with the ground truth locations.

\subsection{Unified Uncertainty-aware Diffusion}

We now present our Unified Uncertainty-aware Diffusion approach, denoted as U2Diff, to infer the set of distributions in \eqref{inference}. Our method can capture the uncertainty associated with each predicted agent state, obtaining both mean and variance of the predicted noise at each denoising step $s$.  

Existing variance-learning approaches in image processing~\cite{nichol2021improved} minimize the variational lower bound by reducing the KL divergence between the predefined true posterior $q(\cdot)$ which follows a scheduled variance $\beta_s$ (see Section \ref{rection_revisiting}), and the model-predicted distribution $p_\theta(\cdot)$. In contrast, our method directly maximizes the likelihood of the noise injected during the forward diffusion pass $\rvepsilon$, by modeling the distribution $\rvepsilon_\theta(\rvepsilon \mid \bmX_s, s, \bmX^\rmo)$ as:
\begin{equation}
 \gN(\rvepsilon; \rvepsilon^{\mu}_\theta(\bmX_s, s, \bmX^\rmo), \rvepsilon^{\sigma}_\theta(\bmX_s, s,\bmX^\rmo)^2\bmI),
\end{equation}
where $\rvepsilon^{\mu}_\theta(\bmX_s, s, \bmX^\rmo)$ and $\rvepsilon^{\sigma}_\theta(\bmX_s, s, \bmX^\rmo)$ are $[T \times N \times 2]$ predicted mean and standard deviation noise, respectively. Particularly, a diagonal covariance matrix across $x$ and $y$ noise components for each agent's state is assumed. 

We propose a novel loss term $\mathcal{L}_\mathrm{NLL}$ that minimizes the Negative Log-Likelihood (NLL) of the noise distribution as:
%\vspace{-3mm}
\begin{equation}
  \mathcal{L}_\mathrm{NLL} = 
  \Scale[0.99]{
  \textrm{log}\left(\sqrt{2 \pi} \rvepsilon^{\sigma}_{\theta}(\bmX_s, s, \bmX_0^\rmo)\right) + 
  \frac{|| \rvepsilon - \rvepsilon^{\mu}_{\theta}(\bmX_s, s, \bmX_0^\rmo) ||_2^2}{2\rvepsilon^{\sigma}_{\theta}(\bmX_s, s, \bmX_0^\rmo)^2} ,}
  \label{eq:loss_nll}
\end{equation}
This regularizer is added to the objective in \eqref{eq:loss}, obtaining the total loss function:
\begin{equation}
  \mathcal{L}_{\mathrm{total}} = \mathcal{L}_{\mathrm{simple}} + \lambda \mathcal{L}_{\mathrm{NLL}},
\end{equation}
where $\lambda$ is a weight factor (typically within the range $0.01$ to $0.001$) that balances the influence of $\mathcal{L}_\mathrm{NLL}$ without overwhelming the primary learning objective. Following the approach in~\cite{nichol2021improved}, we apply a stop-gradient to the predicted noise, $\rvepsilon^{\mu}_{\theta}(\bmX_s, s, \bmX_0^\rmo)$, so that $\mathcal{L}_\mathrm{NLL}$ focuses solely on learning the standard deviation $\rvepsilon^{\sigma}_{\theta}(\bmX_s, s, \bmX_0^\rmo)$. This approach enables the model to represent both expected behavior and the associated uncertainties of agents' trajectories.

\subsubsection{Variance propagation}
During sampling, variance propagation of the predicted noise to the states $(x,y)$ is key. To achieve that, we consider the deterministic Denoising Diffusion Implicit Model (DDIM)~\cite{song2020denoising} with $\zeta$ as the fixed skipping interval: 
\vspace{-2pt}
\begin{equation}
\Scale[0.95]{
  \bmX_{s-\zeta} = \sqrt{\frac{\hat{\alpha}_{s-\zeta}}{\hat{\alpha}_{s}}}\bmX_s + \left(a_{s-\zeta} - \sqrt{\frac{\hat{\alpha}_{s-\zeta}}{\hat{\alpha}_{s}}}a_s \right) \rvepsilon_\theta^\mu(\bmX_s, s, \bmX^\rmo )},%\label{eq:loss_cond}
\end{equation}
where $a_{s} = \sqrt{1 - \hat{\alpha}_s}$. Following the variance rule and similar to~\cite{kou2023bayesdiff}, we approximate the corresponding $\Var(\bmX_{s-\zeta})$ as:
%\vspace{-10pt}
\begin{align}
  & \Scale[1.0]{\frac{\hat{\alpha}_{s-\zeta}}{\hat{\alpha}_s} \Var(\bmX_s) + \left(a_{s-\zeta} - \sqrt{\frac{\hat{\alpha}_{s-\zeta}}{\hat{\alpha}_s}} a_s \right)^2 \rvepsilon^{\sigma}_\theta(\bmX_s, s, \bmX^\rmo)^2   \nonumber} ,
  %\label{eq:loss_cond}
\end{align}
where $\Var(\cdot)$ denotes the variance operator. The variance is initialized as a null tensor in the first denoising step $S$, i.e. $\Var(\bmX_S)=\mathbf{0}$. The covariance term can be approximated as null to avoid high computational cost and potential instabilities without compromising performance.

\begin{figure}[t!]
  \centering
  \includegraphics[width=0.88\linewidth]{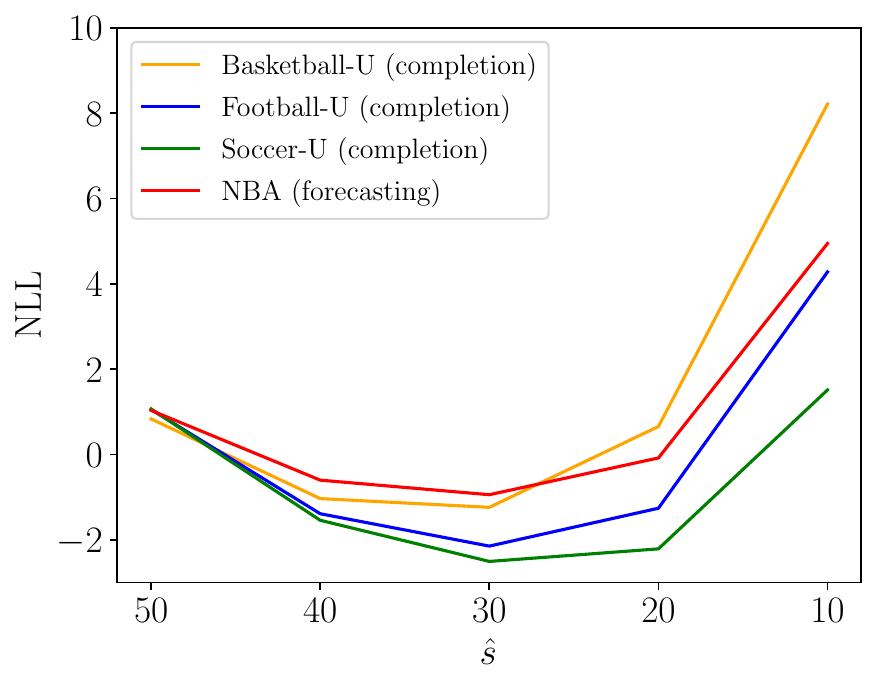}
  \vspace{-4mm}
\caption{\textbf{Evaluation of the NLL over the predicted distribution states} in function of the starting denoising step $\hat{s}$ in which the variance starts propagating.}
  \label{fig:varprop}
  %\vspace{-3mm}
\end{figure}

Our analysis further reveals that beginning variance propagation at a smaller denoising step $\hat{s}$ than $S$, yields optimal performance across the datasets. This suggest that assuming $\Var(\bmX_s) = \mathbf{0}$ for $s\in[\hat{s}, S]$ mitigates the effects of limited data expressivity in the early denoising steps, where the model lacks sufficient information to estimate the meaningful variance. Figure~\ref{fig:varprop} presents the NLL of the predicted state distribution as a function of the starting step $\hat{s}$ for variance propagation across the four datasets. With a total of $S=50$ denoising steps and a skip interval of $\zeta=10$ denoising steps, we find that optimal variance propagation consistently starts at diffusion step $\hat{s}=30$ across all datasets, ensuring robust generalization.

%\vspace{-2mm}
\subsubsection{Architecture}

Inspired by~\cite{tashiro2021csdi}, we introduce our architecture for multi-agent trajectory completion, designed to integrate uncertainty estimation seamlessly. To do so, some modifications are incorporated to enhance performance and adapt to our multi-agent domain. We next present the main ingredients in our contribution.

\noindent \textbf{Input embedding.} Initially, the observed trajectories $\bmX^\rmo$ collected in a tensor with dimensions $T \times N \times 2$, which contain zero values $(0,0)$ for unobserved states, are concatenated with the noised sample $\bmX_s$ of the same dimensionally, producing a combined tensor with dimensions $T \times N \times 4$. This tensor is then transformed into an embedding tensor $J$ through a linear layer followed by a ReLU activation, resulting in dimensions $T \times N \times 256$. That embedding is subsequently processed sequentially through two identical residual denoising blocks.

\begin{figure}[t!]
  \centering
  \includegraphics[width=1.0\linewidth]{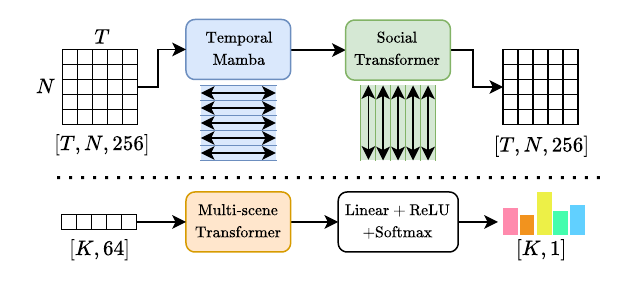}
  \vspace{-5mm}
\caption{\textbf{U2Diff architecture.} \textbf{Top:} Decoupled temporal and social processing in each residual block. \textbf{Bottom:} Multi-scene attention processing and projection with Linear+ReLU+Softmax operations in RankNN to obtain the $K$ error probabilities $e$.}
  \label{fig:architecture}
  \vspace{-3mm}
\end{figure}

\noindent \textbf{Residual denoising block.} Each residual block comprises two main components. First, temporal processing is performed independently for each agent using a bidirectional version of the original MambaSSM~\cite{gu2023mamba}, which we term \textit{Temporal Mamba}. In this step, we compute two separate embeddings for each agent coming from the forward and reverse pass through the MambaSSM. These two embeddings are then summed to capture both past and future temporal information. Second, social processing is conducted to capture interactions between agents at each timestep using a Transformer Encoder~\cite{vaswani2017attention, devlin2018bert}, termed the \textit{Social Transformer}. This decoupling of temporal and social processing is illustrated in Fig.~\ref{fig:architecture}-top, giving the ability to infer scenes with variable timesteps and agents without fixing the temporal and social dimensions. 

Within each residual block, the binary mask $\mathbf{M}$ is used to specify observed and unobserved states, facilitating accurate trajectory completion. Each block outputs a refined tensor $J$ with the same dimensions as its input, along with a skip-connection output $J_\text{skip}$, which is stored from each of the two blocks for later use. Additional details on the implementation of these residual blocks are provided in the supplementary material.

\noindent \textbf{Output tensor.} The output tensor is derived by summing the skip-connection outputs from the two residual blocks, resulting in a tensor of dimensions $T \times N \times 256$. This tensor is then passed through a linear layer with ReLU activation, producing a tensor of shape $T \times N \times 4$. Finally, that result is split to produce the predicted noise, $\rvepsilon^{\mu}_{\theta}(\bmX_s, s, \bmX_0^\rmo)$, while the other tensor component is passed through a sigmoid function to generate the standard deviation $\rvepsilon^{\sigma}_{\theta}(\bmX_s, s, \bmX_0^\rmo)$ with each value bounded in $(0, 1)$.

\subsection{Rank Neural Network}
To compute scene-level uncertainty, we use a simple approach which averages the predicted standard deviations $\sqrt{\Var({\bmX_0)}}$ across all agents and timesteps in a scene. We denote this operation as AvgUcty. 

For a given scene prior observations and its $K$ generated modes, ideally the set of AvgUcty per-mode values and the set of corresponding their error values would correlate positively. In other words, higher AvgUcty values correlate with higher error values. The chosen scene-level metric for the error is the Scene Average Displacement Error (SADE) which is defined as:
\begin{equation}
\label{eq:sade}
    \text{SADE} =\frac{\sum_{n=1}^{N} \sum_{t=1}^{T} \left\lVert \mathbf{\hat{x}}^{n}_{t} - \mathbf{x}^{n}_{t} \right\rVert_2 (1 - \mathbf{m}_{t}^{n})}{\sum_{n=1}^{N} \sum_{t=1}^{T} (1 - \mathbf{m}_{t}^{n})}  \,,
\end{equation}
where $\mathbf{\hat{x}}^{n}_{t}$ and $\mathbf{x}^{n}_{t}$ are the estimation and the corresponding ground truth, respectively, and $\mathbf{m}^{n}_t$ is the value of $\mathbf{M}$ indicating if the $n$-th agent at timestep $t$ is observed or not. 

While AvgUcty provides a straightforward estimation of scene-level uncertainty, it may not fully capture its correlation with SADE. To address this, we introduce a novel learning-based approach that assigns an error probability score, $e$, to each mode, summing to 1 across the $K$ modes. These error probabilities are expected to show a stronger correlation with SADE values compared to AvgUcty. Specifically, we propose a RankNN model, which takes the $K$ generated modes, along with their predicted means and variances, and outputs logits that align with the SADE. 

The objective function to maximize is the Spearman correlation coefficient ($\rho$) between the SADE values and the estimated $e$ values across all $K$ modes. This coefficient evaluates the monotonic relationship between these two sets. Let $e^k$ and SADE$^k$ represent the error probability estimation and the SADE, respectively, for the mode $k\in \{1, \dots, K\}$. This coefficient, defined as the Pearson correlation between rank variables, is computed by first converting each pair ($e^k$, SADE$^k$) for all $K$ modes into differentiable ranks, denoted ($R$[$e^k$], $R$[SADE$^k$]), with $R$[·] being the differentiable rank operator~\cite{blondel2020fast}. Therefore we can express $\rho$ as:
\begin{equation}
\label{eq:spearman}
 \rho = \frac{1}{K} \sum_{k=1}^K \left( \frac{(R[e^k] - \overline{R[e]}) \cdot (R[\text{SADE}^k] - \overline{R[\text{SADE}}])}{\| R[e^k] - \overline{R[e]} \| \cdot \| R[\text{SADE}^k] - \overline{R[\text{SADE}]} \|} \right), \nonumber
\end{equation}
where the terms $\overline{R[e]}$ and $\overline{R[\text{SADE}}]$ are the mean values over the $K$ generated modes. This correlation the normalization in the denominator ensures that is bounded within the interval $(-1, 1)$.

\subsubsection{Architecture}
The architecture takes as input the mean $\bmX_0$ concatenated with its variance $\Var(\bmX_0)$ for each state, creating a $K \times T \times N \times 4$ tensor. This is then extended to $K \times T \times N \times 5$ by appending the binary mask $\mathbf{M}$, repeated $K$ times.

The resulting tensor is embedded to a dimension of 64 and processed through a Temporal Mamba block to capture individual agent dynamics, with operations repeated across $K \times N$. After that, a Social Transformer models social interactions for each timestep, performing operations across $K \times T$. After temporal and social processing, the tensor with dimensions $K \times T \times N \times 64$ is averaged across the timesteps and agents axis, resulting in scene-level embedding tensor $K \times 64$. This tensor is then passed through a Transformer Encoder to perform attention operation across the $K$ scenes, facilitating an efficient ranking. We refer this operation as \textit{Multi-scene Transformer} and is depicted in Fig.~\ref{fig:architecture}-bottom. Finally, a linear layer with ReLU activation produces a vector of length $K$, which is normalized with a softmax function to yield the error probabilities $\{e^1, \dots, e^K\}$. Notably, like our U2Diff, this architecture is flexible as it does not require a fixed number of timesteps $T$, agents $N$, or generated modes $K$ (see supplementary).

\section{Experimental results}

\subsection{Datasets}
For trajectory completion, we evaluate on three team sports datasets~\cite{xu2025sportstraj}: Basketball-U, Football-U, and Soccer-U. \textbf{Basketball-U} derives from NBA dataset~\cite{zhan2018generating} with 93,490 training and 11,543 testing sequences, each spanning 50 frames (8 seconds) capturing $(x,y)$ coordinates for 10 players and the ball. \textbf{Football-U}, based on the NFL Big-Data-Bowl\footnote{https://github.com/nfl-football-ops/Big-Data-Bowl} dataset, contains 10,762 training and 2,624 testing sequences of 50 frames, tracking $(x,y)$ locations for 22 players and the ball. \textbf{Soccer-U}, built from SoccerTrack\footnote{https://github.com/AtomScott/SportsLabKit} dataset, provides 9,882 training and 2,448 testing sequences, each also 50 frames, recording $(x,y)$ positions for 22 players and the ball. In~\cite{xu2025sportstraj}, five masking strategies are defined for trajectory completion, including forecasting futures, imputing in-between states, and inferring the state of over five fully unobserved agents.

For trajectory forecasting, we use the NBA SportVU dataset (\textbf{NBA})~\cite{monti2021dag}, with the same splits and normalization procedure as in LED~\cite{mao2023leapfrog}. The dataset records 30 frames (6 seconds) of $(x,y)$ positions for 10 players and the ball. The prediction task is to observe the first 2 seconds (10 frames) and forecast the subsequent 4 seconds (20 frames).

\subsection{Implementation details}
In our U2Diff, we use $S=50$ diffusion steps during training, with $\lambda$ values set to $0.001$ for Basketball-U dataset and $0.01$ for the other three datasets. The diffusion noise scheduler starts with an initial value of $\beta_0=10^{-4}$ and ends with $\beta_S=0.5$, following a quadratic function. Sampling is performed using DDIM, with a fixed skipping interval of $\zeta=10$ denoising steps, reducing the reverse process to only six denoising steps: $s \in \{50, 40, 30, 20, 10, 1\}$. Optimal variance propagation starts at $\hat{s}=30$. The final step ($s=1$) follows the standard DDPM sampling and the variance is set as $\Var(\bmX_0)=\Var(\bmX_1)$. The Temporal Mamba’s forward and reverse blocks are configured with a hidden size of 256, matching the configuration of the Social Transformer, which uses a 1024-dimensional feedforward layer and 8 attention heads. RankNN training involves generating $20$ modes per scene online using the trained U2Diff with frozen weights. These generated samples are used to compute rankings based on their proximity to ground truth values. Additional implementation details are provided in the supplementary material.

\subsection{Evaluation metrics}
The first set of metrics are the commonly used the agent-wise metrics: minADE$_{K}$ as the minimum average displacement error, and minFDE$_{K}$ as the minimum final displacement error, both calculated over $K$ generated agent-modes. However, these metrics focus only on individual agent modes, lacking a full assessment of inter-agent scene dynamics. To address that, we include scene-level metrics: minSADE$_{K}$ as the minimum SADE (see~\eqref{eq:sade}), and minSFDE$_{K}$ as minimum scene final displacement error, both calculated over $K$ generated scene-modes~\cite{ngiam2021scene, girgis2021latent}.

%Metrica opcional
We also adopt the metric used by~\cite{xu2025sportstraj} in trajectory completion evaluation, here referred to as minADE$_{K}$\cite{xu2025sportstraj}. The Spearman correlation coefficient $\rho$ (see~\eqref{eq:spearman}) is used to assess AvgUcty and RankNN operations. Finally, the Accuracy Rate (AccRate) metric evaluates uncertainty quality by measuring the percentage of ground-truth states that fall within the predicted distribution with 95\% confidence.

\subsection{Comparison in trajectory modeling}
In this section, we compare our approach with several state-of-the-art methods in trajectory completion and trajectory forecasting tasks. 

In Table~\ref{tab:imputation}, we present the results for the minADE$_{20}$\cite{xu2025sportstraj} and, for UniTraj~\cite{xu2025sportstraj} and our baselines, the minSADE$_{20}$ metric (shown in parentheses). Our method outperforms UniTraj, the strongest competing method, across all three completion datasets, achieving over 31\% and 42\% improvements in minADE$_{20}$\cite{xu2025sportstraj} on the Football-U and Soccer-U datasets, respectively. When the number of agents is reduced, as in Basketball-U, we also obtain superior results, with a 27\% improvement in minSADE$_{20}$. The table further includes an ablation study analyzing the impact of loss augmentation ($\lambda = 0$). The results indicate that omitting the loss augmentation does not degrade performance in terms of minSADE$_{20}$. Moreover, when evaluated using the minADE$_{20}$\cite{xu2025sportstraj} metric, loss augmentation leads to improvements across all three datasets.

\begin{table}
\centering
\scalebox{0.68}{
\begin{tabular}{lccc}
    \toprule
    Method & \multicolumn{1}{c}{Basketball-U (Feet)} & \multicolumn{1}{c}{Football-U (Yards)} & \multicolumn{1}{c}{Soccer-U (Pixels)} \\
    \midrule
    Mean & 14.58 & 14.18 & 417.68 \\
    Median & 14.56 & 14.23 & 418.06 \\
    Linear Fit & 13.54 & 12.66 & 398.34 \\   
    LSTM~\cite{hochreiter1997long} & 7.10 & 7.20 & 186.93 \\
    Transformer~\cite{vaswani2017attention} & 6.71 & 6.84 & 170.94 \\
    MAT~\cite{zhan2018generating} & 6.68 & 6.36 & 170.46 \\
    Naomi~\cite{liu2019naomi} & 6.52 & 6.77 & 145.20 \\ 
    INAM~\cite{qi2020imitative} & 6.53 & 5.80 & 134.86 \\
    SSSD~\cite{alcaraz2022diffusion} & 6.18 & 5.08 & 118.71 \\
    GC-VRNN~\cite{xu2023uncovering} & 5.81 & 4.95 & 105.87 \\
    UniTraj~\cite{xu2025sportstraj} & 4.77 (4.29) & 3.55 (4.03) & 94.59 (100.48) \\
    \midrule
     \textbf{U2Diff} ($\lambda = 0$) & \underline{4.68} (\textbf{3.10}) & \underline{2.53} (\underline{2.37}) & \underline{54.41} (\underline{51.27}) \\
    \textbf{U2Diff} & \textbf{4.65} / (\underline{3.13}) & \textbf{2.42} (\textbf{2.35}) & \textbf{53.93} (\textbf{51.14}) \\
    \bottomrule
\end{tabular}
}
\vspace{-2mm}
\caption{\textbf{Evaluation in trajectory completion.} We compare our U2Diff with baseline methods in trajectory completion across three datasets. The metrics used are minADE$_{20}$\cite{xu2025sportstraj}$\downarrow$, and in parentheses, we report the minSADE$_{20}$$\downarrow$ for both our U2Diff and the UniTraj baseline, computed using their original code and publicly available trained model.}
\label{tab:imputation}
\vspace{-2mm}
\end{table}

\begin{table}
\centering
\scalebox{0.68}{
\begin{tabular}{lccc}
    \toprule
    \multirow{2}{*}{Method} & \multicolumn{2}{c}{NBA (Meters)} \\
    \cmidrule(lr){2-3}
    & minADE$_{20}$ / minFDE$_{20}$ $\downarrow$ & minSADE$_{20}$ / minSFDE$_{20}$ $\downarrow$ \\
    \midrule
    MemoNet~\cite{xu2022remember}  & 1.15 / 1.57  & - \\
    NPSN~\cite{bae2022non}     & 1.25 / 1.47  & - \\
    GroupNet~\cite{xu2022groupnet}$^{\chi}$ & 0.94 / 1.22  & 2.12 / 3.72 \\
    AutoBots~\cite{girgis2021latent}$^{\psi}$ &  1.19 / 1.55 & 1.75 / 2.73  \\
    MID~\cite{gu2022stochastic}   & 0.96 / 1.27  &   - \\
    LED~\cite{mao2023leapfrog}   & \textbf{0.81} / \textbf{1.10}  & 1.63 / 2.99 \\
    \midrule
    \textbf{U2Diff} ($\lambda=0$) & 0.86 / \underline{1.11} &  \underline{1.50} / \underline{2.70} \\
    \textbf{U2Diff} & \underline{0.85} / \underline{1.11} & \textbf{1.48} / \textbf{2.68} \\
    \bottomrule
\end{tabular}
}
\vspace{-1mm}
\caption{\textbf{Evaluation in trajectory forecasting.} We compare our U2Diff with baseline methods in trajectory forecasting on NBA dataset. We report four metrics, two agent-wise and two scene-level metrics. $^{\chi}$ means a new pretrained model from their codebase is used, with better results than the reported in the original work. $^{\psi}$ means trained using their original code.}
\label{tab:forecasting}
%\vspace{-5mm}
\end{table}

For the trajectory forecasting task, Table~\ref{tab:forecasting} presents the NBA dataset results. Our unified approach ranks second in agent-wise metrics minADE$_{20}$/minFDE$_{20}$, while achieving state-of-the-art performance in scene-level metrics minSADE$_{20}$/minSFDE$_{20}$, surpassing the diffusion-based LED~\cite{mao2023leapfrog} method by over 9\%. The sequential latent variable model AutoBots~\cite{girgis2021latent} also delivers competitive minSFDE$_{20}$ results. This table includes the same ablation as in Table~\ref{tab:imputation}, showing improvement with our proposed loss.

\begin{figure}[t!]
  \centering
  \includegraphics[width=1.0\linewidth]{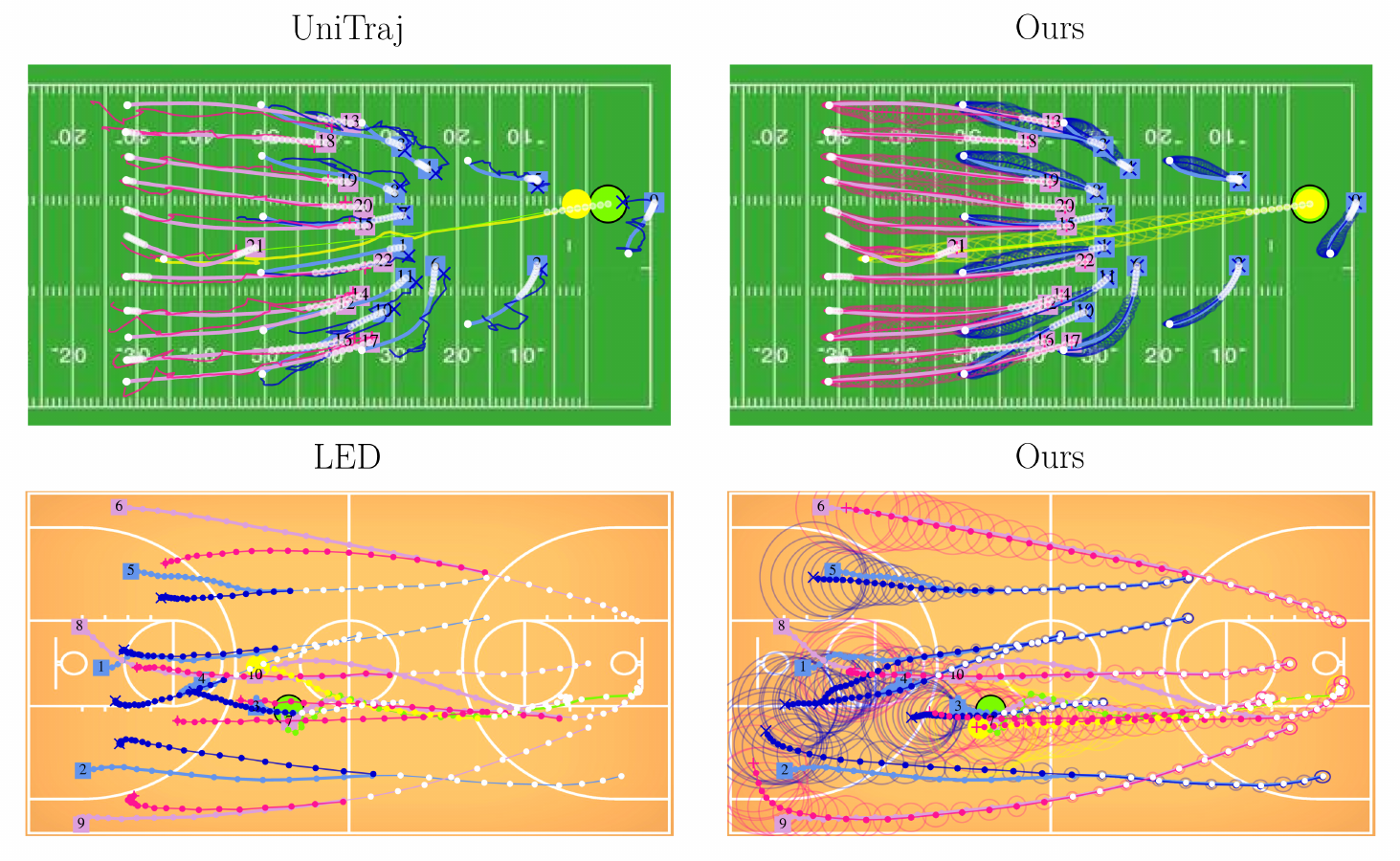}
  \vspace{-6mm}
\caption{\textbf{Qualitative comparisons in trajectory completion (top) and forecasting (bottom).} Our U2Diff is compared with UniTraj~\cite{xu2025sportstraj} for trajectory completion and LED~\cite{mao2023leapfrog} for trajectory forecasting. Ground truth player locations are shown in bright blue and pink, and the ball in green. Model input observations are in white. The predicted mode with the best minSADE$_{20}$ is shown, with players in dark blue and pink, and the ball in yellow.}
  \label{fig:comparison1}
\end{figure}

Figure~\ref{fig:comparison1} illustrates examples for trajectory completion and forecasting. The depicted trajectories are the modes with the minSADE$_{20}$. We compare in tajectory completion against the UniTraj~\cite{xu2025sportstraj}, where our method delivers more accurate reconstructions and plausible predictions. For NBA forecasting, our model shows generally better future predictions—especially for ball trajectories—compared to LED~\cite{mao2023leapfrog}, highlighting the effectiveness of scene-level metrics. Also note that our model is able to estimate variance in both tasks and reconstruct the observed states. Please refer to the supplementary material for additional qualitative results and an ablation study on the U2Diff architecture.

\begin{figure*}
  \centering
  \includegraphics[width=1.0\linewidth]{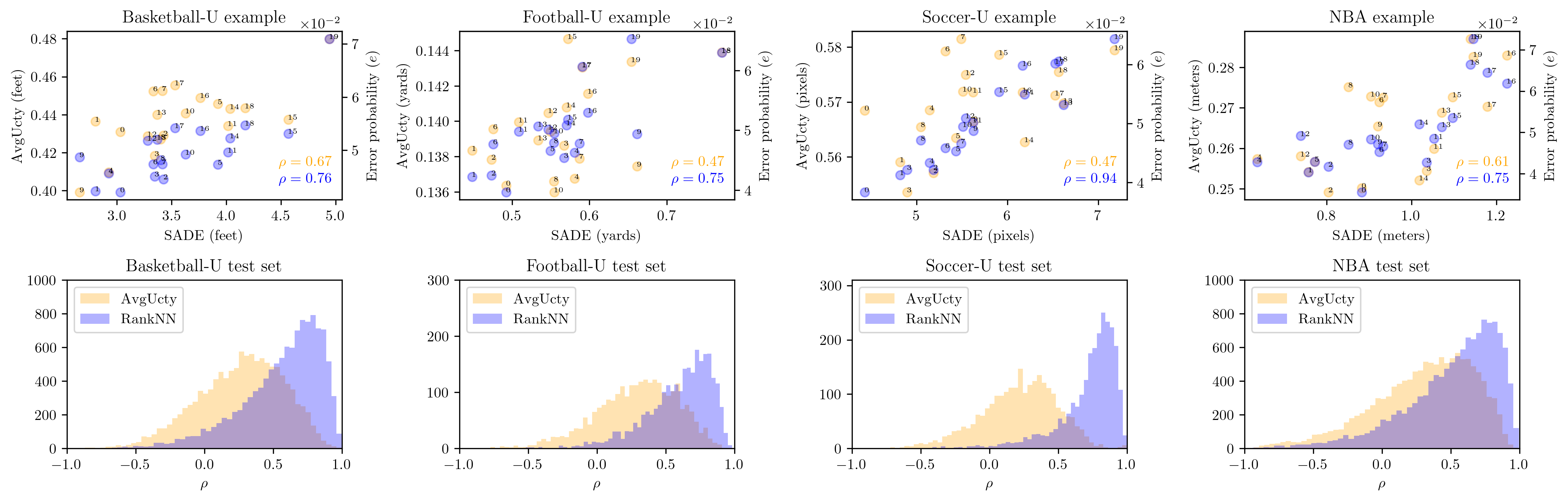}
  \vspace{-8mm}
\caption{\textbf{Qualitative evaluation of the error correlation.} \textbf{Top}: In orange, the AvgUcty versus SADE across the 20 generated modes of a test scene example. In blue, the error probability $e$ versus SADE. \textbf{Bottom}: Distribution of Spearman correlation coefficients $\rho$ for all four test datasets, using AvgUcty in orange and RankNN predicting $e$ in blue.}
  \label{fig:scatter_histograms}
  \vspace{-4mm}
\end{figure*}

\subsection{Uncertainty and error probability estimation}
As previously mentioned, in Fig.~\ref{fig:comparison1} we present the predicted uncertainty for each state based on a 95\% confidence interval. For trajectory completion, the ground-truth states fall within the predicted variance in nearly all cases, indicating robust uncertainty estimation. In contrast, NBA forecasting remains more challenging due to the difficulty in maintaining high confidence over longer prediction horizons.

\begin{table}
\centering
\scalebox{0.68}{
\begin{tabular}{lcccc}
    \toprule
    Sampling &  Basketball-U & Football-U & Soccer-U & NBA  \\
    \midrule
    %           &  0.58   &        &  \\
    Mean              & 82.11 / -1.24 &  92.06 / -2.16 &  94.27 / -2.51  &  76.99 / -0.94 \\
    Top-1 $e$         & 84.01 / -1.41 &  92.82 / -2.24 &  94.75 / -2.57  &  79.19 / -1.03 \\
    Top-1 SADE        & 86.77 / -1.76 &  93.95 / -2.39 &  95.63 / -2.66  &  85.70 / -1.31 \\
    
    \bottomrule
\end{tabular}
}
\vspace{-1mm}
\caption{\textbf{Uncertainty evaluation using the \% of AccRate$\uparrow$ / NLL$\downarrow$ metrics}. Results are depicted for all datasets.}
\label{tab:acc_rate}
%\vspace{-4mm}
\end{table}

To evaluate uncertainty quality, we first use Accuracy Rate (AccRate) and the Negative Log-Likelihood (NLL) in Table~\ref{tab:acc_rate}. We compute these metrics across three sampling strategies, depicted in each row: (Mean) average over all $K$ generated modes, (Top-1 $e$) mode with the lowest error probability $e$ predicted using RankNN, and (Top-1 SADE) mode with the minimum SADE. Football-U and Soccer-U datasets achieve AccRate values over 92\%, indicating strong variance estimation. However, Basketball-U exhibits slightly lower AccRate and higher NLL values, reflecting the increased dynamics of the sport and the challenge of predicting trajectories for five or more fully unseen agents. The NBA dataset, which involves long-horizon forecasting, naturally exhibits lower AccRate and higher NLL. Notably, Top-1 $e$ selection outperforms Mean sampling, achieving higher AccRate and lower NLL, with results approaching Top-1 SADE. This demonstrates the RankNN's effectiveness in identifying reliable modes.

Another key metric is the Spearman correlation $\rho$ between the ranked $K=20$ modes and SADE under the same prior. Ranking can be based on either AvgUcty (predicted uncertainty, relative to $\hat{s}$ in our method) or $e$ (error probabilities). The mean and median values of $\rho$ for each scene across all datasets are shown in Table~\ref{tab:spearmancorr}, with comparisons to AutoBots~\cite{girgis2021latent}. Notably, our predicted uncertainty and the estimated error probabilities achieve higher correlations in the NBA dataset. Note that the AvgUcty operation alone yields moderate correlations when $\hat{s}=30$, with median values ranging from 0.25 to 0.35. This correlation improves further when using the RankNN approach. To illustrate this, Fig.~\ref{fig:scatter_histograms}-top shows four examples where we compare the AvgUcty operation and the error probabilities ($e$) against the SADE for each modes under the same prior. The blue dots corresponding to error probabilities demonstrate better rankings compared to the AvgUcty approach. The distribution of Spearman correlations across the entire test sets is presented in Fig.~\ref{fig:scatter_histograms}-bottom. See supplementary for the ablation analysis of RankNN inputs and components.

\begin{table}
\centering
\scalebox{0.63}{
\begin{tabular}{lcccccc}
    \toprule
    Method &  Rank  & $\hat{s}$ & Basketball-U & Football-U & Soccer-U & NBA  \\
    \midrule
    \multirow{2}{*}{AutoBots~\cite{girgis2021latent}}  & AvgUcty & - &  - & - & - & 0.09 / 0.10  \\
     &  $e$   & - &  - & - & - & 0.37 / 0.44  \\
    \midrule
     \multirow{6}{*}{U2Diff} & \multirow{5}{*}{AvgUcty} & 50 & 0.14 / 0.15 & 0.12 / 0.13 & 0.21 / 0.22 & 0.19 / 0.21 \\  
     & & 40 & 0.22 / 0.24 & 0.25 / 0.26 & 0.22 / 0.24 & 0.30 / 0.35 \\  
     & & 30 & 0.27 / 0.29 & 0.28 / 0.31 & 0.23 / 0.25 & 0.30 / 0.35 \\  
     & & 20 & 0.29 / 0.31 & 0.28 / 0.31 & 0.25 / 0.27 & 0.30 / 0.35 \\  
     & & 10 & 0.29 / 0.31 & 0.26 / 0.28 & 0.25 / 0.28 & 0.29 / 0.33 \\  
     \cmidrule(lr){2-7}
      & $e$   & - &  \textbf{0.56} / \textbf{0.63}   &  \textbf{0.59} / \textbf{0.65}  & \textbf{0.72} / \textbf{0.78} &  \textbf{0.51} / \textbf{0.58}   \\
    \bottomrule
\end{tabular}} 
\vspace{-2mm}
\caption{\textbf{Evaluation of the correlation with error.} The results are the Mean / Median of Spearman correlation ($\rho \uparrow$) between the uncertainty or error probability estimations and the SADE.}
\label{tab:spearmancorr}
\vspace{-3mm}
\end{table}

Finally, Table~\ref{tab:comparison_topk} shows results for different ranking strategies to select the Top-$k$ modes from a set of 20 and then compute the minSADE$_{k}$ for these selected modes. It is important to note that Top-20 is equivalent to minSADE$_{20}$.  The ranking methods considered include Random, AvgUcty (when the model outputs variance) using $\hat{s}=30$ in our framework, and $e$ ranking. We present results for the sampling-based LED~\cite{mao2023leapfrog} generative method as well as AutoBots~\cite{girgis2021latent}, which can infer state-wise uncertainty and error probabilities for the generated modes. Our results show that, by using AvgUcty and the probabilities from RankNN, our method consistently outperforms the others, demonstrating a clear improvement in forecasting accuracy.

\begin{table}
\centering
\scalebox{0.63}{
\begin{tabular}{lcccccc}
    \toprule
    \multirow{2}{*}{Method} & \multirow{2}{*}{Rank} & \multicolumn{5}{c}{NBA (Meters)} \\
    \cmidrule(lr){3-7}
     & &  Top-1  & Top-3  & Top-5  &  Top-10  & Top-20\\
    \midrule
    LED~\cite{mao2023leapfrog} & Random              & 3.80  &  2.17 &  1.92  &  1.73  &  1.63    \\
    \midrule
    \multirow{3}{*}{AutoBots~\cite{girgis2021latent}} & Random        & 2.76  &  2.17 &  2.02  &  1.88  &  1.75\\
     & AvgUcty & 2.37  &  2.19  & 2.09  &  1.94  &  1.75   \\
     & $e$   & 2.40  &  2.08  & 1.98  &  1.86  &  1.75   \\
     \midrule
     \multirow{3}{*}{U2Diff}    & Random          & 2.01  &  1.75  & 1.66 &  1.56  &  1.48     \\
       & AvgUcty &  1.91  &  1.71  & 1.63    & 1.55   & 1.48      \\
      & $e$   &  \textbf{1.82}   &  \textbf{1.66}  &  \textbf{1.60}   & \textbf{1.54}   &  \textbf{1.48}     \\
    \bottomrule
\end{tabular}
}
\vspace{-2mm}
\caption{\textbf{Evaluation of the rank techniques and baselines comparisons.} From 20 generated modes for each method, the Top-$k$ best ones according the ranking method are selected, then the minSADE$_{k}\downarrow$ is computed over this subset.}
\label{tab:comparison_topk}
\vspace{-3mm}
\end{table}

\section{Conclusion}
In this paper, we present U2Diff, a unified uncertainty-aware diffusion framework for general trajectory completion tasks. U2Diff not only outperforms state-of-the-art forecasting baselines in scene-level metrics but also sets a new benchmark in trajectory completion. We demonstrated its effectiveness in estimating state-wise uncertainty via a novel loss augmentation, without sacrificing the accuracy of state predictions. Our experiments reveal that U2Diff's uncertainty estimations exhibit a stronger correlation with ground truth errors compared to the scene-level state-of-the-art method. Additionally, we proposed a novel post-processing supervised RankNN model that infers error probability estimates for each mode, achieving a strong correlation with ground truth errors and also surpassing the scene-level based method.

\noindent \textbf{Acknowledgment.} This work has been supported by the project GRAVATAR PID2023-151184OB-I00 funded by MCIU/AEI/10.13039/501100011033 and by ERDF, UE and by the Government of Catalonia under 2023 DI 00058.

{
    \small
    \bibliographystyle{ieeenat_fullname}
    \bibliography{main}

\begin{thebibliography}{67}
\providecommand{\natexlab}[1]{#1}
\providecommand{\url}[1]{\texttt{#1}}
\expandafter\ifx\csname urlstyle\endcsname\relax
  \providecommand{\doi}[1]{doi: #1}\else
  \providecommand{\doi}{doi: \begingroup \urlstyle{rm}\Url}\fi

\bibitem[Aksan et~al.(2021)Aksan, Kaufmann, Cao, and Hilliges]{aksan2021spatio}
Emre Aksan, Manuel Kaufmann, Peng Cao, and Otmar Hilliges.
\newblock A spatio-temporal transformer for 3d human motion prediction.
\newblock In \emph{2021 International Conference on 3D Vision (3DV)}, pages 565--574. IEEE, 2021.

\bibitem[Alahi et~al.(2016)Alahi, Goel, Ramanathan, Robicquet, Fei-Fei, and Savarese]{alahi2016social}
Alexandre Alahi, Kratarth Goel, Vignesh Ramanathan, Alexandre Robicquet, Li Fei-Fei, and Silvio Savarese.
\newblock Social lstm: Human trajectory prediction in crowded spaces.
\newblock In \emph{Proceedings of the IEEE conference on computer vision and pattern recognition}, pages 961--971, 2016.

\bibitem[Alcaraz and Strodthoff(2022)]{alcaraz2022diffusion}
Juan Miguel~Lopez Alcaraz and Nils Strodthoff.
\newblock Diffusion-based time series imputation and forecasting with structured state space models.
\newblock \emph{arXiv preprint arXiv:2208.09399}, 2022.

\bibitem[Alcorn and Nguyen(2021)]{alcorn2021baller2vec++}
Michael~A Alcorn and Anh Nguyen.
\newblock baller2vec++: A look-ahead multi-entity transformer for modeling coordinated agents.
\newblock \emph{arXiv preprint arXiv:2104.11980}, 2021.

\bibitem[Amirian et~al.(2019)Amirian, Hayet, and Pettr{\'e}]{amirian2019social}
Javad Amirian, Jean-Bernard Hayet, and Julien Pettr{\'e}.
\newblock Social ways: Learning multi-modal distributions of pedestrian trajectories with gans.
\newblock In \emph{Proceedings of the IEEE/CVF Conference on Computer Vision and Pattern Recognition Workshops}, pages 0--0, 2019.

\bibitem[Bae et~al.(2022)Bae, Park, and Jeon]{bae2022non}
Inhwan Bae, Jin-Hwi Park, and Hae-Gon Jeon.
\newblock Non-probability sampling network for stochastic human trajectory prediction.
\newblock In \emph{Proceedings of the IEEE/CVF Conference on Computer Vision and Pattern Recognition}, pages 6477--6487, 2022.

\bibitem[Bae et~al.(2024)Bae, Park, and Jeon]{bae2024singulartrajectory}
Inhwan Bae, Young-Jae Park, and Hae-Gon Jeon.
\newblock Singulartrajectory: Universal trajectory predictor using diffusion model.
\newblock In \emph{Proceedings of the IEEE/CVF Conference on Computer Vision and Pattern Recognition}, pages 17890--17901, 2024.

\bibitem[Blondel et~al.(2020)Blondel, Teboul, Berthet, and Djolonga]{blondel2020fast}
Mathieu Blondel, Olivier Teboul, Quentin Berthet, and Josip Djolonga.
\newblock Fast differentiable sorting and ranking.
\newblock In \emph{International Conference on Machine Learning}, pages 950--959. PMLR, 2020.

\bibitem[Cai et~al.(2020)Cai, Huang, Wang, Cham, Cai, Yuan, Liu, Yang, Zhu, Shen, et~al.]{cai2020learning}
Yujun Cai, Lin Huang, Yiwei Wang, Tat-Jen Cham, Jianfei Cai, Junsong Yuan, Jun Liu, Xu Yang, Yiheng Zhu, Xiaohui Shen, et~al.
\newblock Learning progressive joint propagation for human motion prediction.
\newblock In \emph{Computer Vision--ECCV 2020: 16th European Conference, Glasgow, UK, August 23--28, 2020, Proceedings, Part VII 16}, pages 226--242. Springer, 2020.

\bibitem[Capellera et~al.(2024{\natexlab{a}})Capellera, Ferraz, Rubio, Agudo, and Moreno-Noguer]{capellera2024footbots}
Guillem Capellera, Luis Ferraz, Antonio Rubio, Antonio Agudo, and Francesc Moreno-Noguer.
\newblock Footbots: A transformer-based architecture for motion prediction in soccer.
\newblock In \emph{2024 IEEE International Conference on Image Processing (ICIP)}, pages 2313--2319. IEEE, 2024{\natexlab{a}}.

\bibitem[Capellera et~al.(2024{\natexlab{b}})Capellera, Ferraz, Rubio, Agudo, and Moreno-Noguer]{capellera2024transportmer}
Guillem Capellera, Luis Ferraz, Antonio Rubio, Antonio Agudo, and Francesc Moreno-Noguer.
\newblock Transportmer: A holistic approach to trajectory understanding in multi-agent sports.
\newblock In \emph{Proceedings of the Asian Conference on Computer Vision}, pages 1652--1670, 2024{\natexlab{b}}.

\bibitem[Chai et~al.(2019)Chai, Sapp, Bansal, and Anguelov]{chai2019multipath}
Yuning Chai, Benjamin Sapp, Mayank Bansal, and Dragomir Anguelov.
\newblock Multipath: Multiple probabilistic anchor trajectory hypotheses for behavior prediction.
\newblock \emph{arXiv preprint arXiv:1910.05449}, 2019.

\bibitem[Dendorfer et~al.(2021)Dendorfer, Elflein, and Leal-Taix{\'e}]{dendorfer2021mg}
Patrick Dendorfer, Sven Elflein, and Laura Leal-Taix{\'e}.
\newblock Mg-gan: A multi-generator model preventing out-of-distribution samples in pedestrian trajectory prediction.
\newblock In \emph{Proceedings of the IEEE/CVF International Conference on Computer Vision}, pages 13158--13167, 2021.

\bibitem[Devlin et~al.(2018)Devlin, Chang, Lee, and Toutanova]{devlin2018bert}
Jacob Devlin, Ming-Wei Chang, Kenton Lee, and Kristina Toutanova.
\newblock Bert: Pre-training of deep bidirectional transformers for language understanding.
\newblock \emph{arXiv preprint arXiv:1810.04805}, 2018.

\bibitem[Fang et~al.(2020)Fang, Jiang, Shi, and Zhou]{fang2020tpnet}
Liangji Fang, Qinhong Jiang, Jianping Shi, and Bolei Zhou.
\newblock Tpnet: Trajectory proposal network for motion prediction.
\newblock In \emph{Proceedings of the IEEE/CVF Conference on Computer Vision and Pattern Recognition}, pages 6797--6806, 2020.

\bibitem[Felsen et~al.(2018)Felsen, Lucey, and Ganguly]{felsen2018will}
Panna Felsen, Patrick Lucey, and Sujoy Ganguly.
\newblock Where will they go? predicting fine-grained adversarial multi-agent motion using conditional variational autoencoders.
\newblock In \emph{Proceedings of the European conference on computer vision (ECCV)}, pages 732--747, 2018.

\bibitem[Fragkiadaki et~al.(2015)Fragkiadaki, Levine, Felsen, and Malik]{fragkiadaki2015recurrent}
Katerina Fragkiadaki, Sergey Levine, Panna Felsen, and Jitendra Malik.
\newblock Recurrent network models for human dynamics.
\newblock In \emph{Proceedings of the IEEE international conference on computer vision}, pages 4346--4354, 2015.

\bibitem[Girgis et~al.(2021)Girgis, Golemo, Codevilla, Weiss, D'Souza, Kahou, Heide, and Pal]{girgis2021latent}
Roger Girgis, Florian Golemo, Felipe Codevilla, Martin Weiss, Jim~Aldon D'Souza, Samira~Ebrahimi Kahou, Felix Heide, and Christopher Pal.
\newblock Latent variable sequential set transformers for joint multi-agent motion prediction.
\newblock \emph{arXiv preprint arXiv:2104.00563}, 2021.

\bibitem[Gu and Dao(2023)]{gu2023mamba}
Albert Gu and Tri Dao.
\newblock Mamba: Linear-time sequence modeling with selective state spaces.
\newblock \emph{arXiv preprint arXiv:2312.00752}, 2023.

\bibitem[Gu et~al.(2022)Gu, Chen, Li, Lin, Rao, Zhou, and Lu]{gu2022stochastic}
Tianpei Gu, Guangyi Chen, Junlong Li, Chunze Lin, Yongming Rao, Jie Zhou, and Jiwen Lu.
\newblock Stochastic trajectory prediction via motion indeterminacy diffusion.
\newblock In \emph{Proceedings of the IEEE/CVF Conference on Computer Vision and Pattern Recognition}, pages 17113--17122, 2022.

\bibitem[Guo et~al.(2023)Guo, Du, Shen, Lepetit, Alameda-Pineda, and Moreno-Noguer]{guo2023back}
Wen Guo, Yuming Du, Xi Shen, Vincent Lepetit, Xavier Alameda-Pineda, and Francesc Moreno-Noguer.
\newblock Back to mlp: A simple baseline for human motion prediction.
\newblock In \emph{Proceedings of the IEEE/CVF Winter Conference on Applications of Computer Vision}, pages 4809--4819, 2023.

\bibitem[Gupta et~al.(2018)Gupta, Johnson, Fei-Fei, Savarese, and Alahi]{gupta2018social}
Agrim Gupta, Justin Johnson, Li Fei-Fei, Silvio Savarese, and Alexandre Alahi.
\newblock Social gan: Socially acceptable trajectories with generative adversarial networks.
\newblock In \emph{Proceedings of the IEEE conference on computer vision and pattern recognition}, pages 2255--2264, 2018.

\bibitem[Ho et~al.(2020)Ho, Jain, and Abbeel]{ho2020denoising}
Jonathan Ho, Ajay Jain, and Pieter Abbeel.
\newblock Denoising diffusion probabilistic models.
\newblock \emph{Advances in neural information processing systems}, 33:\penalty0 6840--6851, 2020.

\bibitem[Hochreiter and Schmidhuber(1997)]{hochreiter1997long}
Sepp Hochreiter and J{\"u}rgen Schmidhuber.
\newblock Long short-term memory.
\newblock \emph{Neural computation}, 9\penalty0 (8):\penalty0 1735--1780, 1997.

\bibitem[Hu and Cham(2022)]{hu2022entry}
Bo Hu and Tat-Jen Cham.
\newblock Entry-flipped transformer for inference and prediction of participant behavior.
\newblock In \emph{European Conference on Computer Vision}, pages 439--456. Springer, 2022.

\bibitem[Hu et~al.(2020)Hu, Chen, Zhang, and Gu]{hu2020collaborative}
Yue Hu, Siheng Chen, Ya Zhang, and Xiao Gu.
\newblock Collaborative motion prediction via neural motion message passing.
\newblock In \emph{Proceedings of the IEEE/CVF conference on computer vision and pattern recognition}, pages 6319--6328, 2020.

\bibitem[Jain et~al.(2016)Jain, Zamir, Savarese, and Saxena]{jain2016structural}
Ashesh Jain, Amir~R Zamir, Silvio Savarese, and Ashutosh Saxena.
\newblock Structural-rnn: Deep learning on spatio-temporal graphs.
\newblock In \emph{Proceedings of the ieee conference on computer vision and pattern recognition}, pages 5308--5317, 2016.

\bibitem[Jiang et~al.(2023)Jiang, Cornman, Park, Sapp, Zhou, Anguelov, et~al.]{jiang2023motiondiffuser}
Chiyu Jiang, Andre Cornman, Cheolho Park, Benjamin Sapp, Yin Zhou, Dragomir Anguelov, et~al.
\newblock Motiondiffuser: Controllable multi-agent motion prediction using diffusion.
\newblock In \emph{Proceedings of the IEEE/CVF Conference on Computer Vision and Pattern Recognition}, pages 9644--9653, 2023.

\bibitem[Kim et~al.(2023)Kim, Choi, Kim, Yoon, and Ko]{kim2023ball}
Hyunsung Kim, Han-Jun Choi, Chang~Jo Kim, Jinsung Yoon, and Sang-Ki Ko.
\newblock Ball trajectory inference from multi-agent sports contexts using set transformer and hierarchical bi-lstm.
\newblock \emph{arXiv preprint arXiv:2306.08206}, 2023.

\bibitem[Kosaraju et~al.(2019)Kosaraju, Sadeghian, Mart{\'\i}n-Mart{\'\i}n, Reid, Rezatofighi, and Savarese]{kosaraju2019social}
Vineet Kosaraju, Amir Sadeghian, Roberto Mart{\'\i}n-Mart{\'\i}n, Ian Reid, Hamid Rezatofighi, and Silvio Savarese.
\newblock Social-bigat: Multimodal trajectory forecasting using bicycle-gan and graph attention networks.
\newblock \emph{Advances in Neural Information Processing Systems}, 32, 2019.

\bibitem[Kou et~al.(2023)Kou, Gan, Wang, Li, and Deng]{kou2023bayesdiff}
Siqi Kou, Lei Gan, Dequan Wang, Chongxuan Li, and Zhijie Deng.
\newblock Bayesdiff: Estimating pixel-wise uncertainty in diffusion via bayesian inference.
\newblock \emph{arXiv preprint arXiv:2310.11142}, 2023.

\bibitem[Lee et~al.(2022)Lee, Sohn, Moon, Yoon, Kapadia, and Pavlovic]{lee2022muse}
Mihee Lee, Samuel~S Sohn, Seonghyeon Moon, Sejong Yoon, Mubbasir Kapadia, and Vladimir Pavlovic.
\newblock Muse-vae: Multi-scale vae for environment-aware long term trajectory prediction.
\newblock In \emph{Proceedings of the IEEE/CVF conference on computer vision and pattern recognition}, pages 2221--2230, 2022.

\bibitem[Li et~al.(2023)Li, Li, Ren, Chen, Yuan, and Wang]{li2023bcdiff}
Rongqing Li, Changsheng Li, Dongchun Ren, Guangyi Chen, Ye Yuan, and Guoren Wang.
\newblock Bcdiff: Bidirectional consistent diffusion for instantaneous trajectory prediction.
\newblock \emph{Advances in Neural Information Processing Systems}, 36:\penalty0 14400--14413, 2023.

\bibitem[Liu et~al.(2019)Liu, Yu, Zheng, Zhan, and Yue]{liu2019naomi}
Yukai Liu, Rose Yu, Stephan Zheng, Eric Zhan, and Yisong Yue.
\newblock Naomi: Non-autoregressive multiresolution sequence imputation.
\newblock \emph{Advances in neural information processing systems}, 32, 2019.

\bibitem[Mangalam et~al.(2020)Mangalam, Girase, Agarwal, Lee, Adeli, Malik, and Gaidon]{mangalam2020not}
Karttikeya Mangalam, Harshayu Girase, Shreyas Agarwal, Kuan-Hui Lee, Ehsan Adeli, Jitendra Malik, and Adrien Gaidon.
\newblock It is not the journey but the destination: Endpoint conditioned trajectory prediction.
\newblock In \emph{Computer Vision--ECCV 2020: 16th European Conference, Glasgow, UK, August 23--28, 2020, Proceedings, Part II 16}, pages 759--776. Springer, 2020.

\bibitem[Mao et~al.(2019)Mao, Liu, Salzmann, and Li]{mao2019learning}
Wei Mao, Miaomiao Liu, Mathieu Salzmann, and Hongdong Li.
\newblock Learning trajectory dependencies for human motion prediction.
\newblock In \emph{Proceedings of the IEEE/CVF international conference on computer vision}, pages 9489--9497, 2019.

\bibitem[Mao et~al.(2020)Mao, Liu, and Salzmann]{mao2020history}
Wei Mao, Miaomiao Liu, and Mathieu Salzmann.
\newblock History repeats itself: Human motion prediction via motion attention.
\newblock In \emph{Computer Vision--ECCV 2020: 16th European Conference, Glasgow, UK, August 23--28, 2020, Proceedings, Part XIV 16}, pages 474--489. Springer, 2020.

\bibitem[Mao et~al.(2023)Mao, Xu, Zhu, Chen, and Wang]{mao2023leapfrog}
Weibo Mao, Chenxin Xu, Qi Zhu, Siheng Chen, and Yanfeng Wang.
\newblock Leapfrog diffusion model for stochastic trajectory prediction.
\newblock In \emph{Proceedings of the IEEE/CVF conference on computer vision and pattern recognition}, pages 5517--5526, 2023.

\bibitem[Martinez et~al.(2017)Martinez, Black, and Romero]{martinez2017human}
Julieta Martinez, Michael~J Black, and Javier Romero.
\newblock On human motion prediction using recurrent neural networks.
\newblock In \emph{Proceedings of the IEEE conference on computer vision and pattern recognition}, pages 2891--2900, 2017.

\bibitem[Monti et~al.(2021)Monti, Bertugli, Calderara, and Cucchiara]{monti2021dag}
Alessio Monti, Alessia Bertugli, Simone Calderara, and Rita Cucchiara.
\newblock Dag-net: Double attentive graph neural network for trajectory forecasting.
\newblock In \emph{2020 25th International Conference on Pattern Recognition (ICPR)}, pages 2551--2558. IEEE, 2021.

\bibitem[Navarro and Oh(2022)]{navarro2022social}
Ingrid Navarro and Jean Oh.
\newblock Social-patternn: Socially-aware trajectory prediction guided by motion patterns.
\newblock In \emph{2022 IEEE/RSJ International Conference on Intelligent Robots and Systems (IROS)}, pages 9859--9864. IEEE, 2022.

\bibitem[Ngiam et~al.(2021)Ngiam, Caine, Vasudevan, Zhang, Chiang, Ling, Roelofs, Bewley, Liu, Venugopal, et~al.]{ngiam2021scene}
Jiquan Ngiam, Benjamin Caine, Vijay Vasudevan, Zhengdong Zhang, Hao-Tien~Lewis Chiang, Jeffrey Ling, Rebecca Roelofs, Alex Bewley, Chenxi Liu, Ashish Venugopal, et~al.
\newblock Scene transformer: A unified architecture for predicting multiple agent trajectories.
\newblock \emph{arXiv preprint arXiv:2106.08417}, 2021.

\bibitem[Nichol and Dhariwal(2021)]{nichol2021improved}
Alexander~Quinn Nichol and Prafulla Dhariwal.
\newblock Improved denoising diffusion probabilistic models.
\newblock In \emph{International conference on machine learning}, pages 8162--8171. PMLR, 2021.

\bibitem[Omidshafiei et~al.(2022)Omidshafiei, Hennes, Garnelo, Wang, Recasens, Tarassov, Yang, Elie, Connor, Muller, et~al.]{omidshafiei2022multiagent}
Shayegan Omidshafiei, Daniel Hennes, Marta Garnelo, Zhe Wang, Adria Recasens, Eugene Tarassov, Yi Yang, Romuald Elie, Jerome~T Connor, Paul Muller, et~al.
\newblock Multiagent off-screen behavior prediction in football.
\newblock \emph{Scientific reports}, 12\penalty0 (1):\penalty0 8638, 2022.

\bibitem[Peral et~al.(2025)Peral, Capellera, Rubio, Ferraz, Moreno-Noguer, and Agudo]{peraltemporally}
Marc Peral, Guillem Capellera, Antonio Rubio, Luis Ferraz, Francesc Moreno-Noguer, and Antonio Agudo.
\newblock Temporally accurate events detection through ball possessor recognition in soccer.
\newblock In \emph{Proceedings of the International conference on Computer Vision Theory and Applications}, 2025.

\bibitem[Phan-Minh et~al.(2020)Phan-Minh, Grigore, Boulton, Beijbom, and Wolff]{phan2020covernet}
Tung Phan-Minh, Elena~Corina Grigore, Freddy~A Boulton, Oscar Beijbom, and Eric~M Wolff.
\newblock Covernet: Multimodal behavior prediction using trajectory sets.
\newblock In \emph{Proceedings of the IEEE/CVF conference on computer vision and pattern recognition}, pages 14074--14083, 2020.

\bibitem[Qi et~al.(2020)Qi, Qin, Wu, and Yang]{qi2020imitative}
Mengshi Qi, Jie Qin, Yu Wu, and Yi Yang.
\newblock Imitative non-autoregressive modeling for trajectory forecasting and imputation.
\newblock In \emph{Proceedings of the IEEE/CVF Conference on Computer Vision and Pattern Recognition}, pages 12736--12745, 2020.

\bibitem[Rempe et~al.(2023)Rempe, Luo, Bin~Peng, Yuan, Kitani, Kreis, Fidler, and Litany]{rempe2023trace}
Davis Rempe, Zhengyi Luo, Xue Bin~Peng, Ye Yuan, Kris Kitani, Karsten Kreis, Sanja Fidler, and Or Litany.
\newblock Trace and pace: Controllable pedestrian animation via guided trajectory diffusion.
\newblock In \emph{Proceedings of the IEEE/CVF Conference on Computer Vision and Pattern Recognition}, pages 13756--13766, 2023.

\bibitem[Saadatnejad et~al.(2023)Saadatnejad, Gao, Messaoud, and Alahi]{saadatnejad2023social}
Saeed Saadatnejad, Yang Gao, Kaouther Messaoud, and Alexandre Alahi.
\newblock Social-transmotion: Promptable human trajectory prediction.
\newblock \emph{arXiv preprint arXiv:2312.16168}, 2023.

\bibitem[Sadeghian et~al.(2019)Sadeghian, Kosaraju, Sadeghian, Hirose, Rezatofighi, and Savarese]{sadeghian2019sophie}
Amir Sadeghian, Vineet Kosaraju, Ali Sadeghian, Noriaki Hirose, Hamid Rezatofighi, and Silvio Savarese.
\newblock Sophie: An attentive gan for predicting paths compliant to social and physical constraints.
\newblock In \emph{Proceedings of the IEEE/CVF conference on computer vision and pattern recognition}, pages 1349--1358, 2019.

\bibitem[Salzmann et~al.(2020)Salzmann, Ivanovic, Chakravarty, and Pavone]{salzmann2020trajectron++}
Tim Salzmann, Boris Ivanovic, Punarjay Chakravarty, and Marco Pavone.
\newblock Trajectron++: Multi-agent generative trajectory forecasting with heterogeneous data for control.
\newblock \emph{arXiv preprint arXiv:2001.03093}, 2, 2020.

\bibitem[Shi et~al.(2023)Shi, Wang, Zhou, and Hua]{shi2023trajectory}
Liushuai Shi, Le Wang, Sanping Zhou, and Gang Hua.
\newblock Trajectory unified transformer for pedestrian trajectory prediction.
\newblock In \emph{Proceedings of the IEEE/CVF International Conference on Computer Vision}, pages 9675--9684, 2023.

\bibitem[Song et~al.(2020)Song, Meng, and Ermon]{song2020denoising}
Jiaming Song, Chenlin Meng, and Stefano Ermon.
\newblock Denoising diffusion implicit models.
\newblock \emph{arXiv preprint arXiv:2010.02502}, 2020.

\bibitem[Sun et~al.(2019)Sun, Karlsson, Wu, Tenenbaum, and Murphy]{sun2019stochastic}
Chen Sun, Per Karlsson, Jiajun Wu, Joshua~B Tenenbaum, and Kevin Murphy.
\newblock Stochastic prediction of multi-agent interactions from partial observations.
\newblock \emph{arXiv preprint arXiv:1902.09641}, 2019.

\bibitem[Tashiro et~al.(2021)Tashiro, Song, Song, and Ermon]{tashiro2021csdi}
Yusuke Tashiro, Jiaming Song, Yang Song, and Stefano Ermon.
\newblock Csdi: Conditional score-based diffusion models for probabilistic time series imputation.
\newblock \emph{Advances in Neural Information Processing Systems}, 34:\penalty0 24804--24816, 2021.

\bibitem[Vaswani et~al.(2017)Vaswani, Shazeer, Parmar, Uszkoreit, Jones, Gomez, Kaiser, and Polosukhin]{vaswani2017attention}
Ashish Vaswani, Noam Shazeer, Niki Parmar, Jakob Uszkoreit, Llion Jones, Aidan~N Gomez, {\L}ukasz Kaiser, and Illia Polosukhin.
\newblock Attention is all you need.
\newblock \emph{Advances in neural information processing systems}, 30, 2017.

\bibitem[Xu et~al.(2022{\natexlab{a}})Xu, Li, Ni, Zhang, and Chen]{xu2022groupnet}
Chenxin Xu, Maosen Li, Zhenyang Ni, Ya Zhang, and Siheng Chen.
\newblock Groupnet: Multiscale hypergraph neural networks for trajectory prediction with relational reasoning.
\newblock In \emph{Proceedings of the IEEE/CVF Conference on Computer Vision and Pattern Recognition}, pages 6498--6507, 2022{\natexlab{a}}.

\bibitem[Xu et~al.(2022{\natexlab{b}})Xu, Mao, Zhang, and Chen]{xu2022remember}
Chenxin Xu, Weibo Mao, Wenjun Zhang, and Siheng Chen.
\newblock Remember intentions: Retrospective-memory-based trajectory prediction.
\newblock In \emph{Proceedings of the IEEE/CVF Conference on Computer Vision and Pattern Recognition}, pages 6488--6497, 2022{\natexlab{b}}.

\bibitem[Xu et~al.(2023{\natexlab{a}})Xu, Tan, Tan, Chen, Wang, Wang, and Wang]{xu2023eqmotion}
Chenxin Xu, Robby~T Tan, Yuhong Tan, Siheng Chen, Yu~Guang Wang, Xinchao Wang, and Yanfeng Wang.
\newblock Eqmotion: Equivariant multi-agent motion prediction with invariant interaction reasoning.
\newblock In \emph{Proceedings of the IEEE/CVF Conference on Computer Vision and Pattern Recognition}, pages 1410--1420, 2023{\natexlab{a}}.

\bibitem[Xu and Fu(2025)]{xu2025sportstraj}
Yi Xu and Yun Fu.
\newblock Sports-traj: A unified trajectory generation model for multi-agent movement in sports.
\newblock In \emph{The Thirteenth International Conference on Learning Representations}, 2025.

\bibitem[Xu et~al.(2023{\natexlab{b}})Xu, Bazarjani, Chi, Choi, and Fu]{xu2023uncovering}
Yi Xu, Armin Bazarjani, Hyung-gun Chi, Chiho Choi, and Yun Fu.
\newblock Uncovering the missing pattern: Unified framework towards trajectory imputation and prediction.
\newblock In \emph{Proceedings of the IEEE/CVF Conference on Computer Vision and Pattern Recognition}, pages 9632--9643, 2023{\natexlab{b}}.

\bibitem[Yang et~al.(2024)Yang, Su, Gkanatsios, Ke, Jain, Schneider, and Fragkiadaki]{yang2024diffusion}
Brian Yang, Huangyuan Su, Nikolaos Gkanatsios, Tsung-Wei Ke, Ayush Jain, Jeff Schneider, and Katerina Fragkiadaki.
\newblock Diffusion-es: Gradient-free planning with diffusion for autonomous driving and zero-shot instruction following.
\newblock \emph{arXiv preprint arXiv:2402.06559}, 2024.

\bibitem[Yeh et~al.(2019)Yeh, Schwing, Huang, and Murphy]{yeh2019diverse}
Raymond~A Yeh, Alexander~G Schwing, Jonathan Huang, and Kevin Murphy.
\newblock Diverse generation for multi-agent sports games.
\newblock In \emph{Proceedings of the IEEE/CVF Conference on Computer Vision and Pattern Recognition}, pages 4610--4619, 2019.

\bibitem[Yuan et~al.(2021)Yuan, Weng, Ou, and Kitani]{yuan2021agentformer}
Ye Yuan, Xinshuo Weng, Yanglan Ou, and Kris~M Kitani.
\newblock Agentformer: Agent-aware transformers for socio-temporal multi-agent forecasting.
\newblock In \emph{Proceedings of the IEEE/CVF International Conference on Computer Vision}, pages 9813--9823, 2021.

\bibitem[Zhan et~al.(2018)Zhan, Zheng, Yue, Sha, and Lucey]{zhan2018generating}
Eric Zhan, Stephan Zheng, Yisong Yue, Long Sha, and Patrick Lucey.
\newblock Generating multi-agent trajectories using programmatic weak supervision.
\newblock \emph{arXiv preprint arXiv:1803.07612}, 2018.

\bibitem[Zhao et~al.(2021)Zhao, Gao, Lan, Sun, Sapp, Varadarajan, Shen, Shen, Chai, Schmid, et~al.]{zhao2021tnt}
Hang Zhao, Jiyang Gao, Tian Lan, Chen Sun, Ben Sapp, Balakrishnan Varadarajan, Yue Shen, Yi Shen, Yuning Chai, Cordelia Schmid, et~al.
\newblock Tnt: Target-driven trajectory prediction.
\newblock In \emph{Conference on Robot Learning}, pages 895--904. PMLR, 2021.

\bibitem[Zheng et~al.(2016)Zheng, Yue, and Hobbs]{zheng2016generating}
Stephan Zheng, Yisong Yue, and Jennifer Hobbs.
\newblock Generating long-term trajectories using deep hierarchical networks.
\newblock \emph{Advances in Neural Information Processing Systems}, 29, 2016.

\end{thebibliography}
}

% WARNING: do not forget to delete the supplementary pages from your submission 
% \input{sec/X_suppl}

\end{document}

\maketitle
\thispagestyle{empty}

\section{Unified Uncertainty-Aware Diffusion}
In this section we depict the details of our Unified Uncertainty-aware Diffusion (U2Diff) approach.

\subsection{Implementation}
U2Diff is trained for 100 epochs on an RTX A6000, with a batch size of 64 for the Basketball-U dataset and 16 for the others. Training time varies between 30 to 120 minutes, depending on the dataset size. For the forecasting task in NBA dataset (see Table 2 of the main paper), individual-based metrics minADE$_{20}$/minFDE$_{20}$ are computed using directly DDPM sampling. The remaining other cases are computed using the DDIM, as explained in the main paper.

\subsection{Architecture details}
\begin{figure*}
  \centering
  \includegraphics[trim={1cm 0.1cm 3.5cm 0}, clip, width=1.0\linewidth]{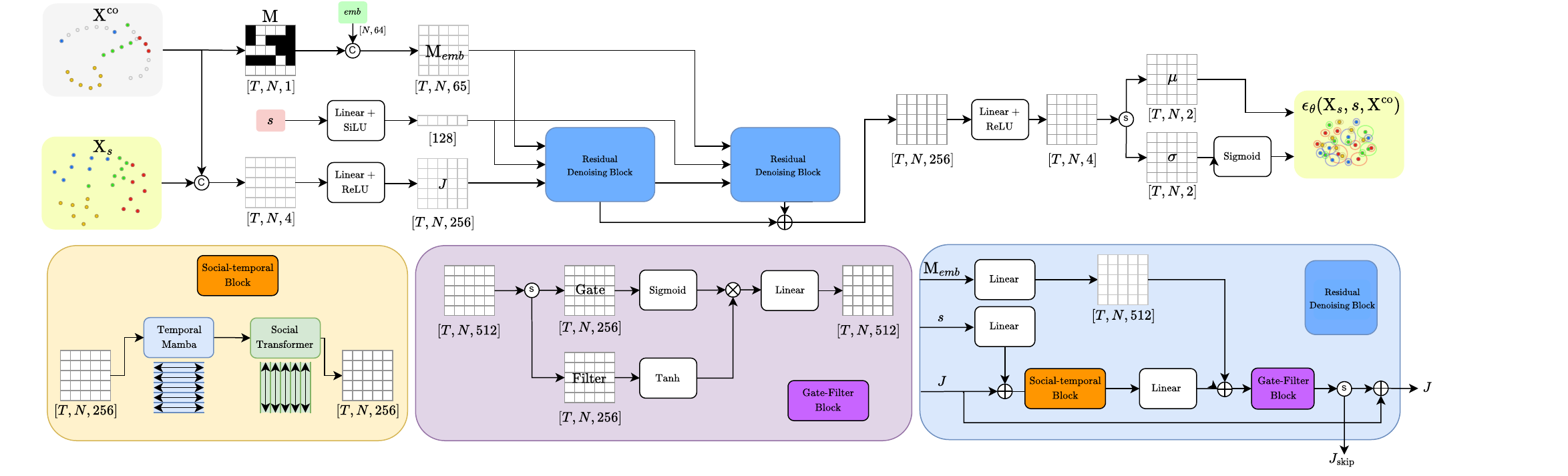}
  \vspace{-4mm}
\caption{\textbf{Unified Uncertainty-aware Diffusion Model (U2Diff)}}
  \label{fig:u2diff_arch}
  \vspace{-3mm}
\end{figure*}

The proposed architecture builds upon CSDI framework [55], introducing key modifications tailored to the task of multi-agent trajectory completion. The first modification replaces the original Transformer-based temporal processing module with the Temporal Mamba, specifically designed to model temporal dynamics without the need for explicit positional encodings. This change addresses the sub-optimal performance of temporal positional encodings observed during the denoising process. Additionally, the final linear head of the architecture is redesigned to output both the mean and standard deviation of the predicted noise distribution, enabling uncertainty estimation. The full architecture is depicted in Fig.~\ref{fig:u2diff_arch}.

\noindent \textbf{Input Embedding} As outlined in the main paper, the input embedding tensor $J$ of dimension $T \times N \times 256$ is constructed from the concatenation of tensors $\bmX^\rmo$ and $\bmX_s$. The denoising step index $s$ is also embedded into a 128-dimensional space using a linear layer followed by a SiLU activation function before being incorporated into each residual denoising block. Additionally, a learnable embedding of dimension 64 is assigned for each agent, referred to as $emb$, and concatenated with the binary mask $\mathbf{M}$, forming the mask/embedding tensor, denoted as $\mathbf{M}_{emb}$.

\noindent \textbf{Residual Denoising Block} Each residual denoising block processes three inputs: the embedding tensor $J$, the embedded denoising step $s$, and the mask/embedding tensor $\mathbf{M}_{emb}$. First, the embedding of $s$ is projected to a 256-dimensional space using a linear transformation to align with the dimensionality of $J$. This transformed embedding is then summed to each state in $J$. Next, the resulting tensor passes through the \textbf{Social-temporal Block} (see Fig.~\ref{fig:u2diff_arch}), which sequentially applies the Temporal Mamba and the Social Transformer.
The output tensor from this block is projected to a 512-dimensional space. In parallel, the tensor $\mathbf{M}_{emb}$ is also embedded into the same $T \times N \times 512$ dimensional space. These two tensors are summed to form the input to the next stage.

This combined tensor is processed by a \textbf{Gate-filter Block}. Here the tensor is split into two components: $\text{Gate}$ and $\text{Filter}$, both of size $T \times N \times 256$. The $\text{Gate}$ is passed through a sigmoid activation, while the $\text{Filter}$ is passed through a hyperbolic tangent (Tanh). The two resulting tensors are state-wise multiplied (Hadamard product) and then projected back into a 512-dimensional space via a linear transformation. The output of the Gate-filter Block is split into two tensors of size $T \times N \times 256$. The first tensor is added to the original $J$, producing a refined version of $J$ that is passed to the next residual denoising block. The second tensor serves as the skip connection output, $J_\text{skip}$, of the current residual denoising block.

%ACABAR DE EXPLICAR EL GATE-FILTER.

\noindent \textbf{Output tensor} The $J_\text{skip}$ outputs from all residual denoising blocks are summed together and passed through a linear layer followed by a ReLU activation function. This operation produces a tensor of dimension $T \times N \times 4$. This tensor is further split into two tensor of size $T \times N \times 2$ each: predicted mean noise, representing the central tendency of the noise distribution $\rvepsilon^{\mu}_\theta(\bmX_s, s, \bmX^\rmo)$, and the predicted standard deviation $\rvepsilon^{\sigma}_\theta(\bmX_s, s, \bmX^\rmo)$, obtained after applying a Sigmoid function to bound the values within the interval (0,1). Both tensors are symbolized as $\mu$ and $\sigma$ in Fig.\ref{fig:u2diff_arch}.

This design ensures that the model not only reconstructs trajectories but also quantifies uncertainty at each predicted state by modeling the noise distribution with a diagonal covariance structure, as explained in the main paper.

\subsection{Architecture ablation}

The main paper provides an ablation analysis on setting $\lambda=0$ or training from scratch, demonstrating the influence of the proposed loss function. Here, we present a detailed ablation study focusing on the key architectural components of U2Diff to further substantiate their contributions.

Table~\ref{tab:ablation_u2diff} reports results using the minSADE$_{20}$ metric across all datasets for the following configurations: \textbf{w/o TM} (without the Temporal Mamba) which is replaced with a Transformer Encoder, reverting to as the original design in CSDI [55], and \textbf{w/o ST} (without Social Transformer) responsible for encoding agent interactions.

We showcase the importance of replacing Temporal Mamba by transformers to enhance significantly the performance with respect the original CSDI [55]. Also we show the extreme importance in encoding the agents interactions with the Social Transformer in our multi-agent domain.

\begin{table}[h]
\centering
\scalebox{0.75}{
\begin{tabular}{lcccc}
    \toprule
     \multirow{2}{*}{U2Diff}   &  Basketball-U & Football-U & Soccer-U & NBA  \\
     &  (Feet) & (Yards) & (Pixels) & (Meters)  \\
    \midrule
    w/o TM (CSDI [55])   &  3.74  & 2.70  & 58.78  &  1.73   \\
    w/o ST       &  4.06 & 4.75 & 96.66  &  1.92   \\
    \midrule
    Ours ($\lambda = 0$) & \textbf{3.10}  & 2.37  & 51.27 & 1.50 \\
    Ours        & 3.13  & \textbf{2.35}  & \textbf{51.14}  &  \textbf{1.48}  \\
    
    \bottomrule
\end{tabular}
}
\vspace{-1mm}
\caption{\textbf{Ablation study on U2Diff using the minSADE$_{20}\downarrow$ metric}. The results are presented across all evaluated datasets.}
\label{tab:ablation_u2diff}
\vspace{-4mm}
\end{table}

\section{Rank Neural Network}
This section provides an in-depth explanation of the Rank Neural Network (RankNN) approach.

\subsection{Implementation}
During training, we generate $K=20$ modes online using the U2Diff model with trained and frozen weights. The RankNN architecture is trained using a batch size of 32 scenes for 20 epochs. Depending on the dataset, each epoch on a RTX A6000 takes between 30 and 180 minutes. 

To ensure convergence in Basketball-U and Soccer-U datasets, where stochasticity posed challenges, we first pretrained RankNN using online generations from a suboptimal U2Diff model trained with a batch size four times larger. Since the suboptimal model produces more distinguishable variations in quality, this pretraining phase makes it easier for RankNN to learn meaningful ranking patterns before fine-tuning on higher-quality generations.

\subsection{Architecture details}
\begin{figure*}
  \centering
  \includegraphics[width=0.9\linewidth]{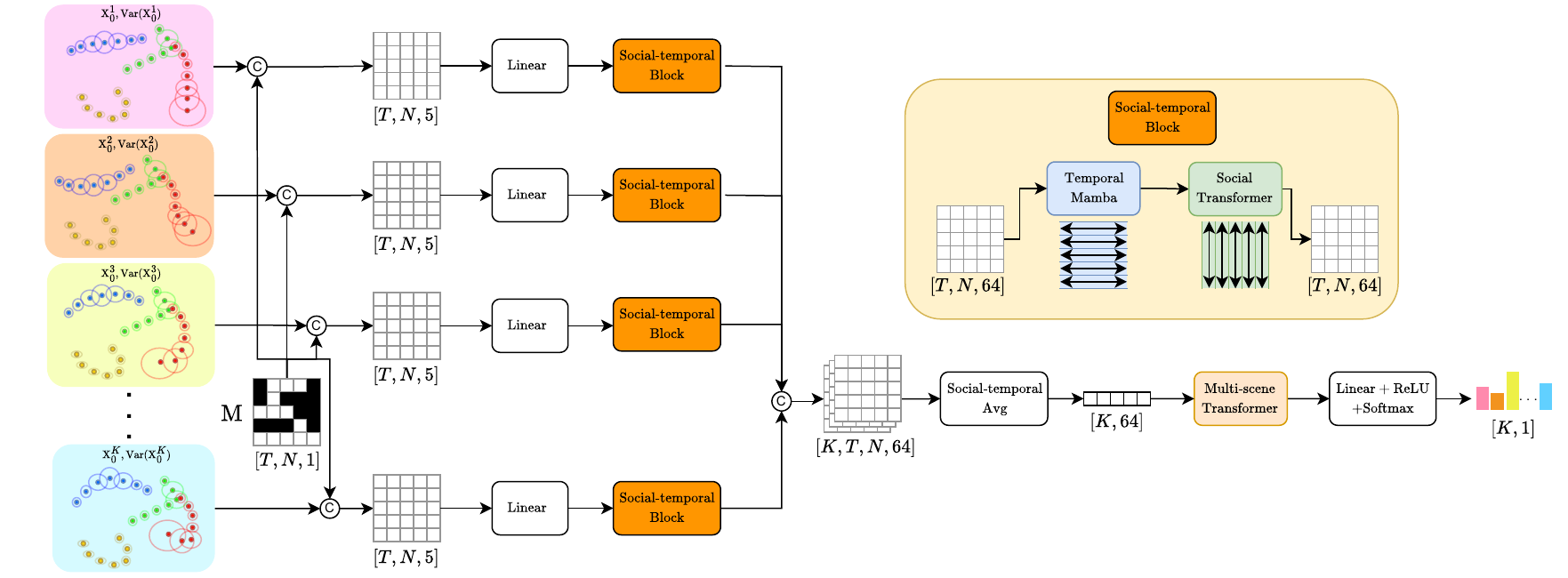}
  \vspace{-4mm}
\caption{\textbf{Rank Neural Network architecture (RankNN)}.}
  \label{fig:ranknn_arch}
  \vspace{-3mm}
\end{figure*}

To complement the explanation provided in the main paper, Fig.~\ref{fig:ranknn_arch} presents a visual diagram of the RankNN architecture. Similar to U2Diff, the Social-Temporal operation involves sequential processing through the Temporal Mamba and Social Transformer modules.

\subsection{Architecture ablation}
We perform an ablation study on the RankNN architecture, presented in Table~\ref{tab:ablation_ranknn}, using the Spearman correlation coefficient ($\rho$) as the evaluation metric to measure the alignment between $e$ and SADE. The study evaluated the following configurations: \textbf{w/o TM w/o ST} skips the Social-temporal Block entirely, bypassing both Temporal Mamba and Social Transformer processing; \textbf{w/o TM} excludes the Temporal Mamba processing, while retaining the Social Transformer; \textbf{w/o ST} excludes the Social Transformer processing, while retaining the Temporal Mamba; \textbf{w/o MST} skips the Multi-scene Transformer processing; and \textbf{w/o VAR} removes the predicted variance $\Var(\bmX_0)$ from the input to RankNN, using only the mean predicted locations locations $\bmX_0$ and the binary mask $\mathbf{M}$. The last configuration reduces the input tensor from $K \times T \times N \times 5$ to $K \times T \times N \times 3$.  

The results emphasize the importance of processing states through the Social-Temporal Block, demonstrating the critical contributions of both the Temporal Mamba and the Social Transformer. Furthermore, the w/o MST configuration results in sub-optimal performance, particularly on the Football-U dataset, underscoring the value of Multi-scene Transformer processing. Finally, the w/o VAR configuration highlights the essential role of the predicted variance $\Var(\bmX_0)$ from U2Diff.

\begin{table}[h]
\centering
\scalebox{0.80}{
\begin{tabular}{lcccc}
    \toprule
     RankNN   &  Basketball-U & Football-U & Soccer-U & NBA  \\
    \midrule
    %           &  0.58   &        &  \\
    w/o TM w/o ST   &  0.28 & 0.37 & 0.31 & 0.40 \\
    w/o TM          &  0.35 & 0.50 & 0.54 & 0.45 \\
    w/o ST          &  0.39 & 0.51 & 0.34 & 0.45 \\
    w/o MST         &  0.52 & 0.47 & 0.65 & 0.50 \\
    w/o VAR         &  0.54 & 0.56 & 0.55  &  0.50 \\  
    \midrule
    Ours            & \textbf{0.56}  &  \textbf{0.59} & \textbf{0.72}  &  \textbf{0.51} \\
    \bottomrule
\end{tabular}
}
\vspace{-1mm}
\caption{\textbf{Ablation study on RankNN using the Mean of Spearman correlation coefficient ($\rho\uparrow$)}. The results are presented across all evaluated datasets.}
\label{tab:ablation_ranknn}
\vspace{-4mm}
\end{table}

\section{Qualitative results}
\begin{figure*}[t!]
  \centering
  \includegraphics[width=1.0\linewidth]{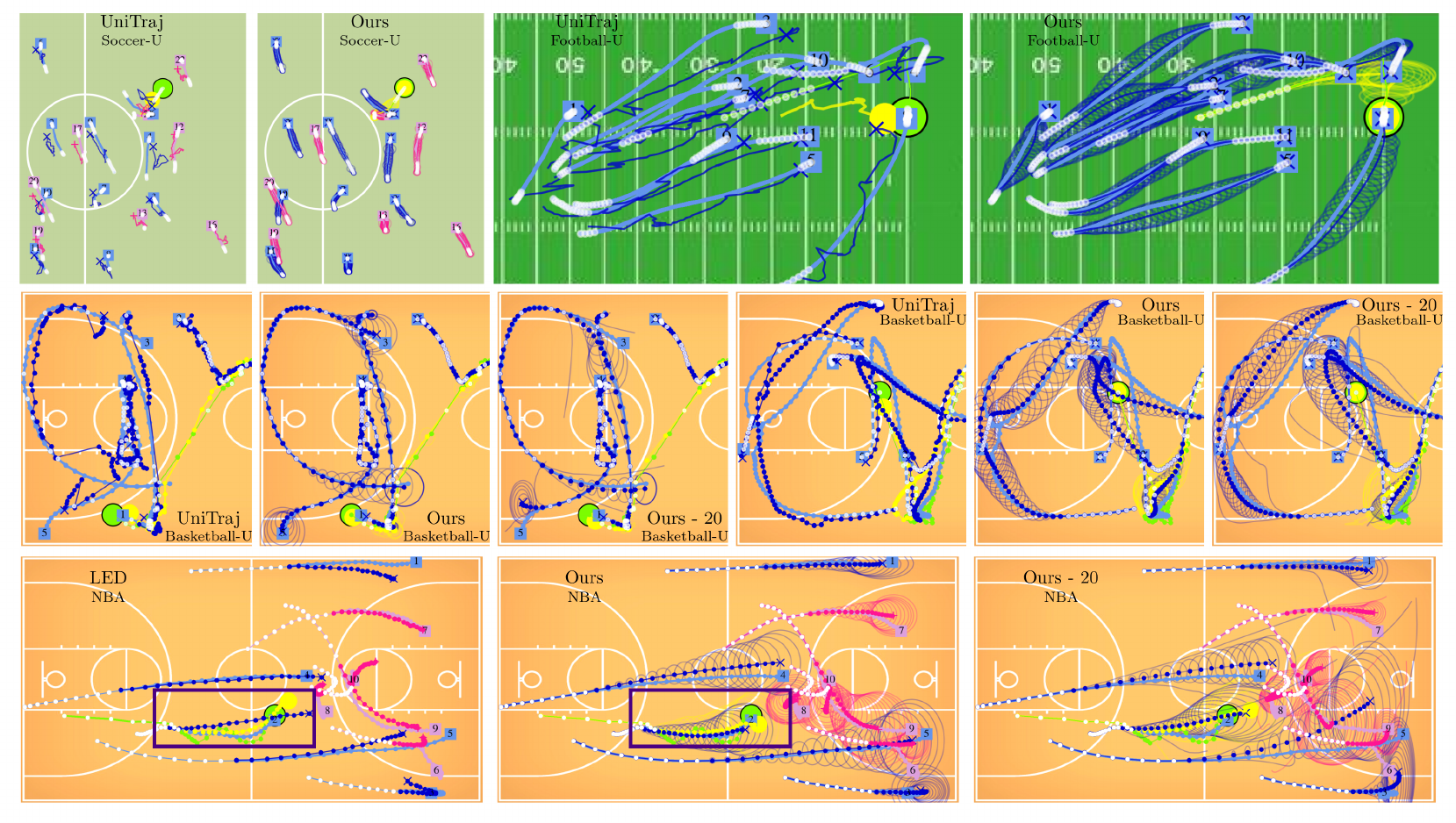}
  \vspace{-8mm}
\caption{\textbf{Qualitative comparisons in trajectory completion and forecasting}. Ground truth player locations are shown in bright blue and pink, and the ball in green. Model input observations are represented in white. In LED, Unitraj and Ours, the predicted mode with the best minSADE$_{20}$ is shown, with players in dark blue and pink, and the ball in yellow. Our model’s predictions also include estimated state-wise variances. In Ours-20, the prediction distribution over 20 samples is illustrated, including upper and lower bound modes.}
  \label{fig:rebut_qual}
  \vspace{-5mm}
\end{figure*}

We present additional qualitative results in Fig.~\ref{fig:rebut_qual}, which extends Figure 4 from the main paper. In this figure, we compare the performance of our method with LED[38] and UniTraj[60] across all four datasets. Notably, our method outperforms LED in capturing the ball-possessor relationship, a key aspect of multi-agent sports scenarios. We explicitly highlight the ball-possessor with a pink rectangle, making it easier to visualize how different methods handle this critical interaction. Regarding the completion tasks, our method shows better alignment of the predictions with the ground truth, especially in the already observed-reconstructed states, compared UniTraj.

In addition, we include the variant Ours-20, which represents the distribution of predictions over 20 sampled trajectories. This variant serves to highlight the alignment between the predicted variance from our model (Ours) and the sampling approximation (Ours-20), demonstrating how our approach effectively captures uncertainty in trajectory predictions.

